\documentclass{article}

\usepackage{microtype}
\usepackage{graphicx}
\usepackage{subfigure}
\usepackage{booktabs} % for professional tables

\usepackage{amsmath}
\usepackage{amssymb}
\usepackage{xcolor}
\usepackage{tikz-cd}

\usepackage{hyperref}

\newcommand{\rotscene}{R^{\mathcal{Z}}}
\newcommand{\rotrender}{R^{\mathcal{X}}}
\newcommand{\rotdeltascene}{R_{\boldsymbol{\theta}}^{\mathcal{Z}}}
\newcommand{\rotdeltarender}{R_{\boldsymbol{\theta}}^{\mathcal{X}}}

\definecolor{darkgray1}{rgb}{0.66, 0.66, 0.66}
\definecolor{darkgreen}{HTML}{2CA02C}

\usepackage[accepted]{icml2020}

\icmltitlerunning{Equivariant Neural Rendering}

\begin{document}

\twocolumn[
\icmltitle{Equivariant Neural Rendering}



\icmlsetsymbol{equal}{*}

\begin{icmlauthorlist}
\icmlauthor{Emilien Dupont}{oxford}
\icmlauthor{Miguel Angel Bautista}{apple}
\icmlauthor{Alex Colburn}{apple}
\icmlauthor{Aditya Sankar}{apple}
\icmlauthor{Carlos Guestrin}{apple}  % REMOVE THIS WHEN DOING ARXIV VERSION
\icmlauthor{Joshua Susskind}{apple}
\icmlauthor{Qi Shan}{apple}
\end{icmlauthorlist}

\icmlaffiliation{oxford}{University of Oxford, UK}
\icmlaffiliation{apple}{Apple Inc, USA}

\icmlcorrespondingauthor{Emilien Dupont}{dupont@stats.ox.ac.uk}
\icmlcorrespondingauthor{Qi Shan}{qshan@apple.com}

\icmlkeywords{Machine Learning, ICML}

\vskip 0.3in
]



\printAffiliationsAndNotice{}  % leave blank if no need to mention equal contribution

\begin{abstract}
We propose a framework for learning neural scene representations directly from images, without 3D supervision. Our key insight is that 3D structure can be imposed by ensuring that the learned representation \textit{transforms} like a real 3D scene. Specifically, we introduce a loss which enforces equivariance of the scene representation with respect to 3D transformations. Our formulation allows us to infer and render scenes in real time while achieving comparable results to models requiring minutes for inference. In addition, we introduce two challenging new datasets for scene representation and neural rendering, including scenes with complex lighting and backgrounds. Through experiments, we show that our model achieves compelling results on these datasets as well as on standard ShapeNet benchmarks.

%Unlike most neural rendering frameworks, we make no assumptions about the rendering process.
%Our model has extremely weak assumptions, it does not require absolute coordinates and pose at inference time. Further, we can infer and render scenes in real time whereas most other methods require minutes to perform inference.

\end{abstract}

\section{Introduction}

Designing useful 3D scene representations for neural networks is a challenging task. While several works have used traditional 3D representations such as voxel grids \cite{maturana2015voxnet, nguyen2018rendernet, zhu2018visual}, meshes \cite{jack2018learning}, point clouds \cite{qi2017pointnet, insafutdinov2018unsupervised} and signed distance functions \cite{DeepSDF}, they each have limitations. For example, it is often difficult to scalably incorporate texture, lighting and background into these representations. Recently, neural scene representations have been proposed to overcome these problems \cite{Eslami1204, sitzmann2019deepvoxels, sitzmann2019scene}, usually by incorporating ideas from graphics rendering into the model architecture.

In this paper, we argue that equivariance with respect to 3D transformations provides a strong inductive bias for neural rendering and scene representations. Indeed, we argue that, for many tasks, scene representations need not be explicit (such as point clouds and meshes) as long as they \textit{transform} like explicit representations.

%\textbf{Weak assumptions} 
%Our formulation allows us to bypass the need for 3D supervision and does not pose any restrictions on the rendering process. As a result, we are able to model complex visual effects such as reflections and cluttered backgrounds, which is not trivial for more restricted rendering functions.

\begin{figure}[t]
\begin{center}

\centerline{
    \raisebox{1.25pt}{\includegraphics[width=0.159\columnwidth]{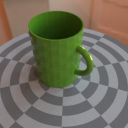}}%
    \hspace{3pt}%
    \textcolor{darkgray1}{\vrule width 1pt height 35pt depth -4pt}%
    \hspace{2pt}%
    \includegraphics[width=0.83\columnwidth]{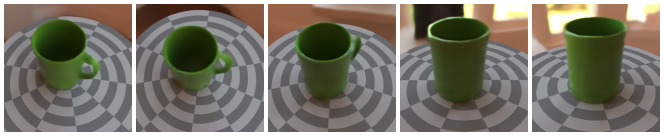}
}

\centerline{
    \raisebox{1.25pt}{\includegraphics[width=0.159\columnwidth]{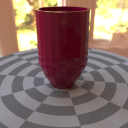}}%
    \hspace{3pt}%
    \textcolor{darkgray1}{\vrule width 1pt height 35pt depth -4pt}%
    \hspace{2pt}%
    \includegraphics[width=0.83\columnwidth]{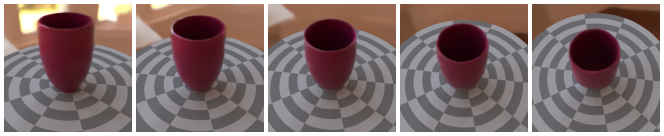}
}
\vspace{-5pt}
\caption{From a single image (left), our model infers a scene representation and generates new views of the scene (right) with a learned neural renderer.}
\label{intro-fig}
\end{center}
\vspace{-20pt}  % REMOVE THIS WHEN DOING ARXIV VERSION
\end{figure}

Our model is trained with no 3D supervision and only requires images and their relative poses to learn equivariant scene representations. Our formulation does not pose any restrictions on the rendering process and, as result, we are able to model complex visual effects such as reflections and cluttered backgrounds. Unlike most other scene representation models \cite{Eslami1204, sitzmann2019deepvoxels, sitzmann2019scene}, our model does not require any pose information at inference time. From a single image, we can infer a scene representation, transform it and render it (see Fig. \ref{intro-fig}). Further, we can infer and render scene representations in real time while many scene representation algorithms require minutes to perform inference from an image or a set of images \cite{nguyen2018rendernet, sitzmann2019scene, DeepSDF}.

% No rendering assumptions (should, in principle, be able to deal with smoke, reflections and other complicated effects).
% Equivariance is a weak assumption (encompasses a larger class of models)
% Purely based on how scenes transform
% Model does not require 3D supervision.
%Single shot novel view/scene inference (most models require more than one image)
%We make no assumptions about the rendering process: this allows us to model complex visual effects like reflections and so on which are not trivial with other methods.
%We show that you don't need to make very restrictive assumptions, equivariance can for many purposes be enough.
%We could imagine our algorithm works for example on data with complicated effects such as e.g. transparency or reflection (or smoke or things like that). Unlike other approaches that try to restrict the decoder to be some kind of renderer like function (e.g. ray tracing or volume rendering) all we do is restrict the transformation properties of the renderer. This still gives it complete freedom to render things that would be otherwise difficult to do. Further, this restriction also means there is a big improvement in performance over neural nets that just directly render image (they tend not to have 3D structure).

%\textbf{Generalization} 
While several works achieve impressive results by training models on images of a single scene and then generating novel views of that same scene \cite{mildenhall2020nerf}, we focus on generalizing across different scenes. This provides an additional challenge as we are required to learn a prior over shapes and textures to generalize to novel scenes. Our approach also allows us to bypass the need for different scenes in the training set to be aligned (or share the same coordinate system). Indeed, since we learn scene representations that \textit{transform} like real scenes, we only require \textit{relative} transformations to train the model. This is particularly advantageous when considering real scenes with complicated backgrounds where alignment can be difficult to achieve.

Neural rendering and scene representation models are usually tested and benchmarked on the ShapeNet dataset \cite{chang2015shapenet}. However, the images produced from this dataset are often very different from real scenes: they are rendered on empty backgrounds and only involve a single fixed object. As our model does not rely on 3D supervision, we are able to train it on rich data where it is very expensive or difficult to obtain 3D ground truths. We therefore introduce two new datasets of posed images which can be used to test models with complex visual effects. The first dataset, MugsHQ, is composed of photorealistic renders of colored mugs on a table with an ambient backgroud. The second dataset, 3D mountains, contains renders of more than 500 mountains in the Alps using satellite and topography data. In summary, our contributions are:

\begin{itemize}
    \item We introduce a framework for learning scene representations and novel view synthesis without explicit 3D supervision, by enforcing equivariance between the change in viewpoint and change in the latent representation of a scene.
    \item We show that we can generate convincing novel views in real time without requiring alignment between scenes nor pose at inference time.
    \item We release two new challenging datasets to test representations and neural rendering for complex, natural scenes, and show compelling rendering results on each, highlighting the versatility of our method.
\end{itemize}

% we argue that equivariance with respect to 3d transformations in combination with spatial representations provide a strong inductive bias for neural rendering and scene representations

%Description of model
%- Challenging to build a model that enforces equivariance
%- We want to build a model that can infer a 3d scene representation from a single image (refer to fig. 1)
%- Requires us to define a new layer

% While the idea of learning equivariant representations is simple, building such models in practice is challenging. We propose a model composed of an inverse renderer mapping an image to a neural scene representation and a forward neural renderer generating images from representations. The scene representations themselves are 3 dimensional tensors which can undergo the same transformations as an explicit 3D scene. In this work, we focus on 3D rotations although the model can be generalized to other symmetry transformations such as translation and scaling. Defining tensor rotations in 3D is not straightforward and we show that naive tensor rotations cannot be used to learn equivariant representations. Instead we propose a new differentiable layer, invertible shear rotation, which allows for this. 

%Advantages of our model
%- infer a scene representation and modify and rerender it from a single image
%- no 3d ground truths are necessary.
%- Representation is coordinate free.
%- Unlike most algorithms, does not require pose at test time.
%- Scene representation in a single forward pass
%- Rerendering in a single forward pass too

\vspace{3pt}
\section{Related Work}

\textbf{Scene representations.} Traditional scene representations (e.g. point clouds, voxel grids and meshes) do not scale well due to memory and compute requirements. Truncated signed distance functions (SDF) have been used to aggregate depth measurements from 3D sensors \cite{curless_tsdf} to map and track surfaces in real-time \cite{newcombe2011kinectfusion}, without requiring assumptions about surface structure. \citeauthor{niessner2013real} extend these implicit surface methods by incrementally fusing depth measurements into a hashed memory structure. More recently \citeauthor{DeepSDF} extend SDF representations to whole classes of shapes, with learned neural mappings. Similar implicit neural representations have been used for 3D reconstruction from a single view \cite{xu2019disn,mescheder2018occupancy}. % Our method can be seen as an implicit neural representation encoded into a latent 3D tensor.

\textbf{Neural rendering.} Neural rendering approaches produce photorealistic renderings given noisy or incomplete 3D or 2D observations. In \citeauthor{DeferredNeuralRendering}, incomplete 3D inputs are converted to rich scene representations using neural textures, which fill in and regularize noisy measurements. \citeauthor{sitzmann2019scene} encode geometry and appearance into a latent code that is decoded using a differentiable ray marching algorithm. Similar to our work, DeepVoxels~\cite{sitzmann2019deepvoxels} encodes scenes into a 3D latent representation. In contrast with our work, these methods either require 3D information during training, complicated rendering priors or expensive inference schemes. % Our model makes very weak assumptions. Contrast this with other models which do not.
% Mention generalization (many models work well but don't generalize).
% Mention that it is very fast. Real time inference and real time rendering.

\textbf{Novel view synthesis.} In \citeauthor{Eslami1204}, one or more input views with camera poses are aggregated into a context feature vector, and are rendered into a target 2D image given a query camera pose. \citeauthor{tobin-egqn} extend this base method using epipolar geometrical constraints to improve the decoding. Our model does not require the expensive sequential decoding steps of these models and enforces 3D structure through equivariance. \citeauthor{tatarchenko2016multi} can perform novel view synthesis for single objects consistent with a training set, but require depth to train the model. \citeauthor{Hedman2018,Hedman2016,thies2018headon,Xu2019} use coarse geometric proxies. Our method only requires images and their poses to train, and can therefore extend more readily to real scenes with minimal assumptions about geometry. Works based on flow estimation for view synthesis \cite{sun2018multi, zhou2016view} predict a flow field over the input image(s) conditioned on a camera viewpoint transformation. These approaches model a free-form deformation in image space, as a result, they cannot explicitly enforce equivariance with respect to 3D rotation. In addition, these models are commonly restricted to single objects, not entire scenes. % these models are commonly restricted to segmented single objects, not entire scenes.
% Mention large angles. Many models work well with small angles (where this is not generative process basically)

\textbf{Equivariance.} While translational equivariance is a natural property of convolution on the spatial grid, traditional neural networks are not equivariant with respect to general transformation groups. Equivariance for discrete rotations can be achieved by replicating and rotating filters \cite{pmlr-v48-cohenc16}. Equivariance to rotation has been extended to 3D using spherical CNNs \cite{esteves17}. Steerable filters \cite{cohen2016steerable} and equivariant capsule networks \cite{NIPS2018_8100} achieve approximate smooth equivariance by estimating pose and transforming filters, or by disentangling pose and filter representations. \citeauthor{worrall2017interpretable} use equivariance to learn autoencoders with interpretable transformations, although they do not explicitly encode 3D structure in the latent space. \citeauthor{olszewski2019transformable}'s method is closely related to ours but only focuses on a limited range of transformations, instead of complete 3D rotations. In our method, we achieve equivariance by treating our latent representation as a geometric 3D data structure and applying rotations directly to this representation.

\vspace{3pt}
\section{Equivariant Scene Representations}

We denote an image by $\mathbf{x} \in \mathcal{X} = \mathbb{R}^{c \times h \times w}$ where $c, h, w$ are the number of channels, height and width of the image respectively. We denote a scene representation by $\mathbf{z} \in \mathcal{Z}$. We further define a rendering function $g : \mathcal{Z} \to \mathcal{X}$ mapping scene representations to images and an inverse renderer $f : \mathcal{X} \to \mathcal{Z}$ mapping images to scenes.

\begin{figure}[t]
\begin{center}
\centerline{\includegraphics[width=0.7\columnwidth]{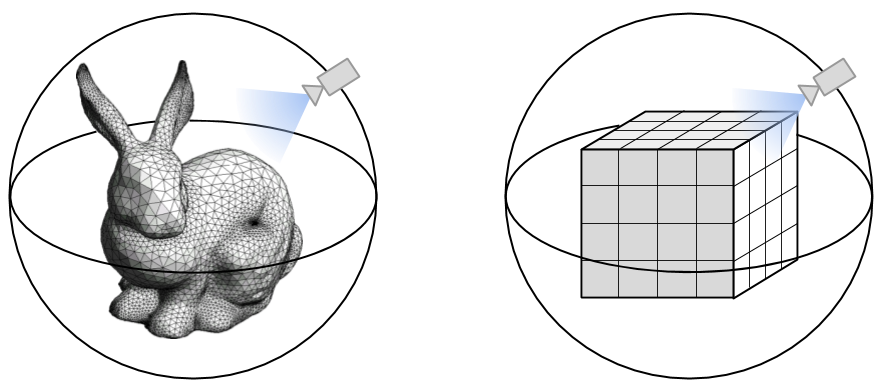}}
\vspace{-7pt}
\caption{Left: A camera on the sphere observing an explicit scene representation (a mesh). Right: A camera on the sphere observing an implicit scene representation (a 3D tensor).}
\label{camera-positions}
\end{center}
\vspace{-18pt}
\end{figure}

We distinguish between two classes of scene representations: explicit and implicit representations (see Fig. \ref{camera-positions}). Explicit representations are designed to be interpreted by humans and are rendered by a fixed interpretable process. As an example, $\mathbf{z}$ can be a 3D mesh and $g$ a standard rendering function such as a raytracer. Implicit representations, in contrast, are abstract and need not be human interpretable. For example, $\mathbf{z}$ could be the latent space of an autoencoder and $g$ a neural network. We argue that, for many tasks, scene representations need not be explicit as long as they \textit{transform} like explicit representations.

\begin{figure}[t]
\begin{center}
    \begin{tikzcd}
        \includegraphics[trim={4cm 4cm 4cm 5cm}, clip, width=0.22\columnwidth]{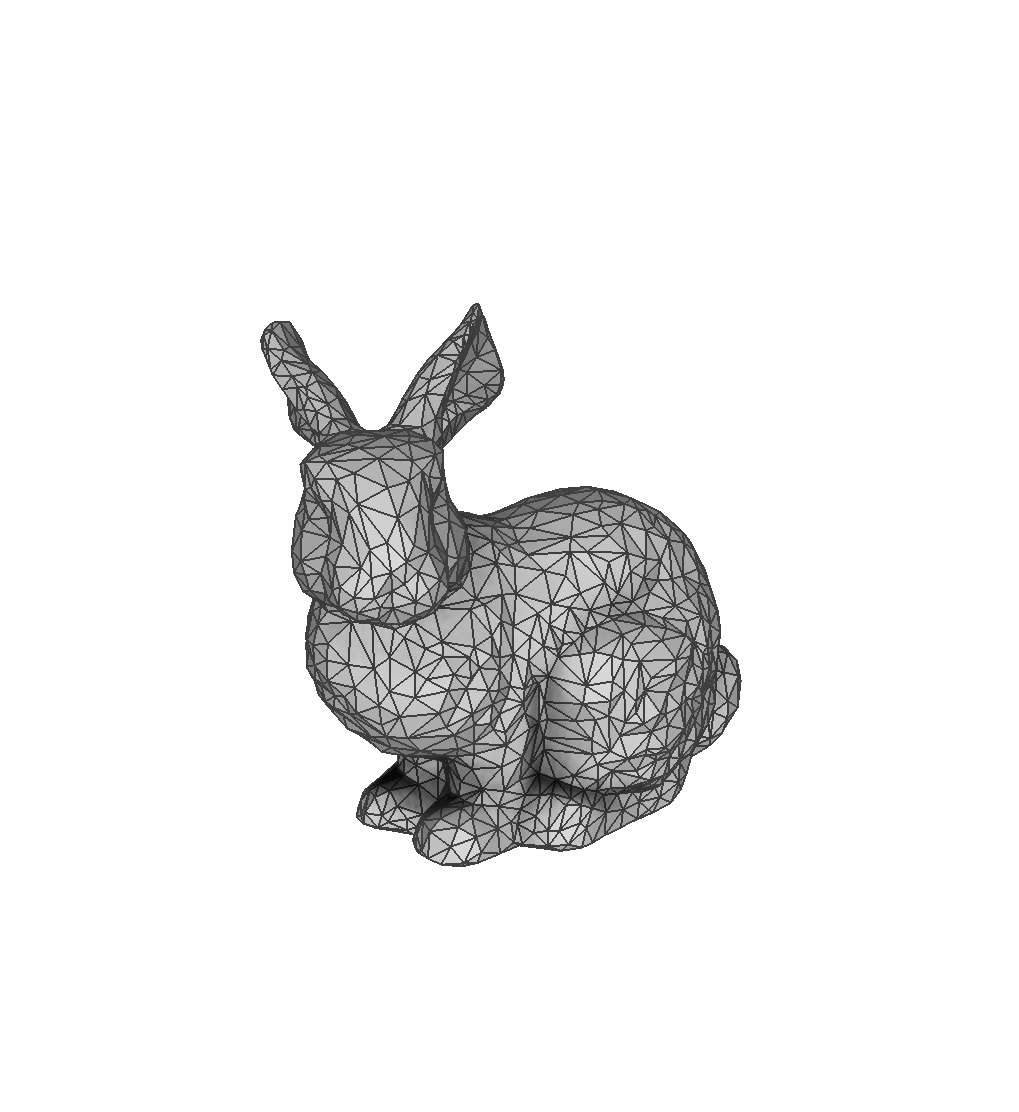} 
        \arrow[r, "\huge T^{\mathcal{Z}}", yshift=20pt]
        \arrow[d, "g"]
        & \includegraphics[trim={4cm 4cm 4cm 5cm}, clip, width=0.22\columnwidth]{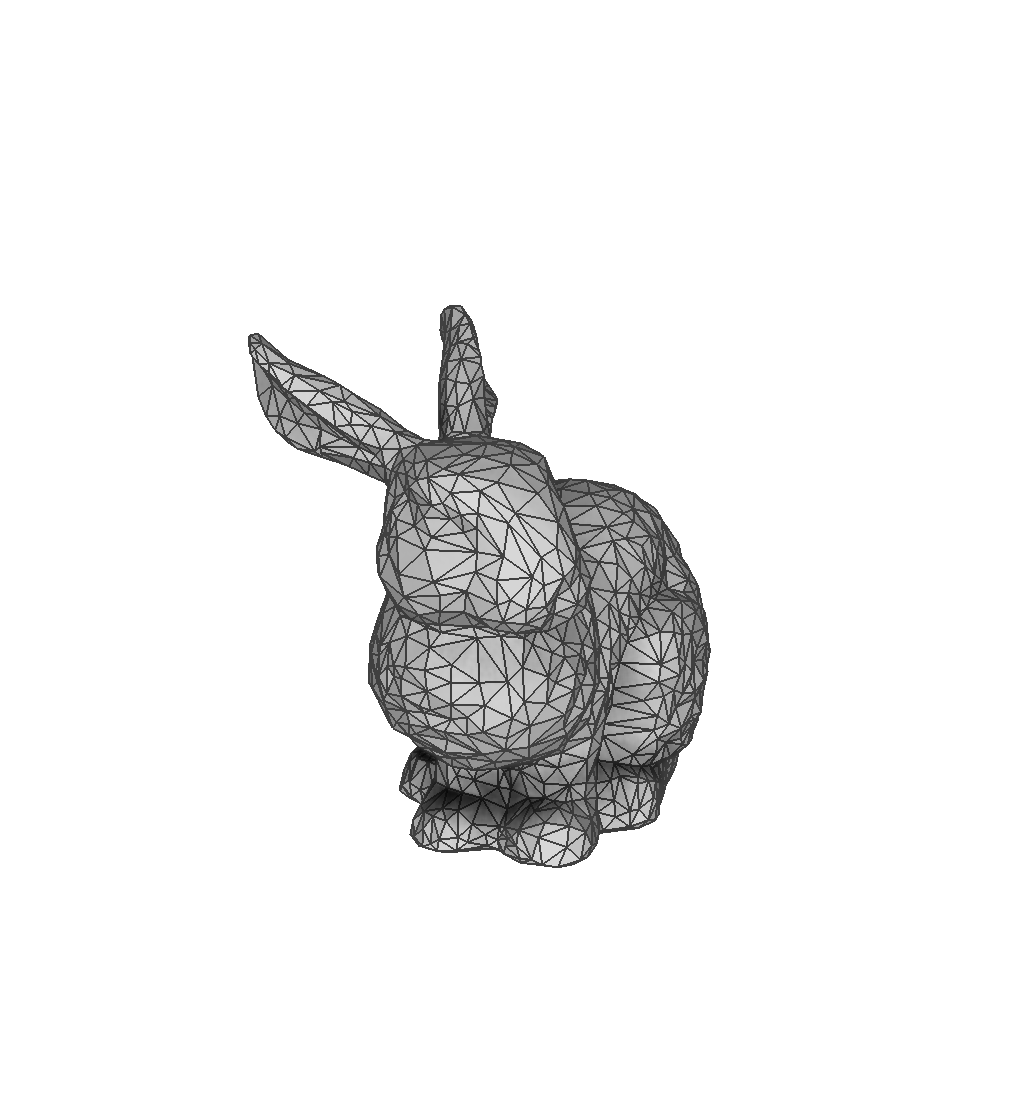} 
        \arrow[d, "g"] \\ 
        \includegraphics[trim={4cm 3cm 4cm 6cm}, clip, width=0.25\columnwidth]{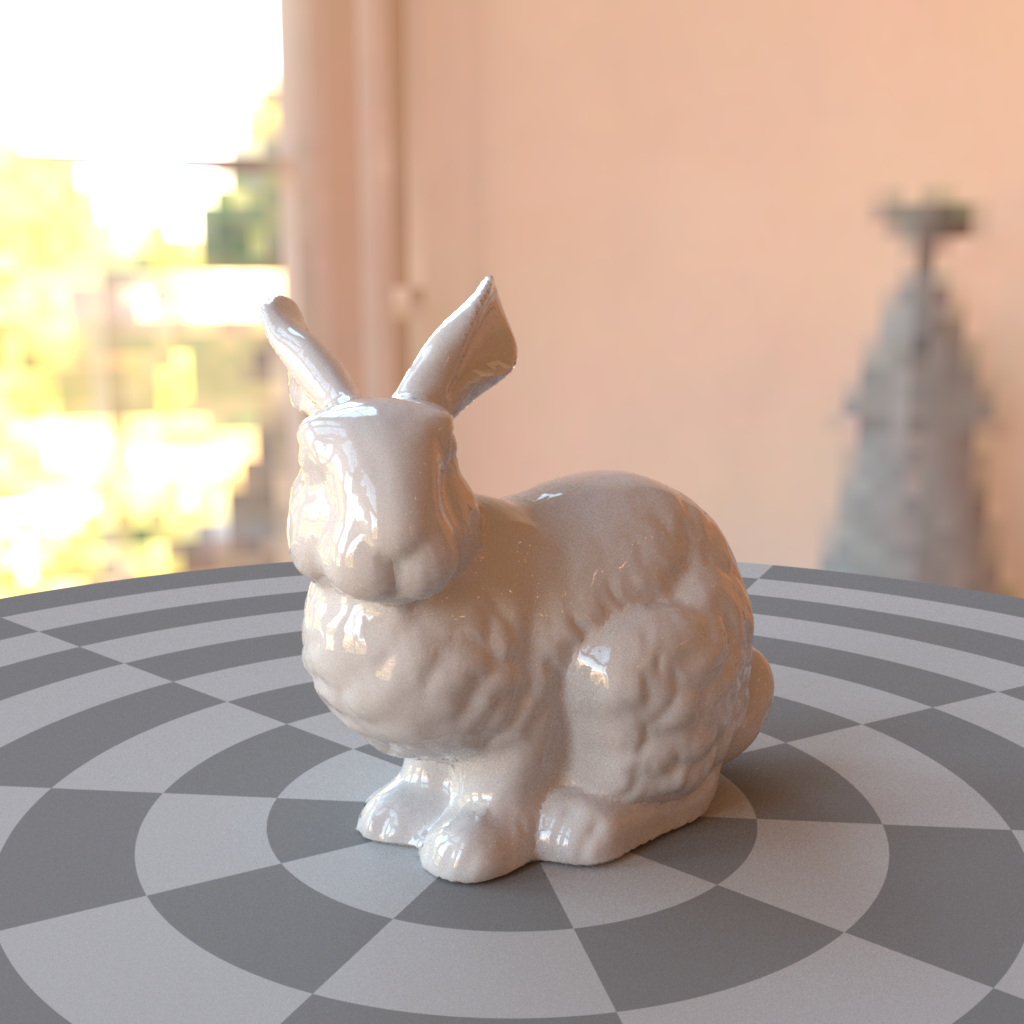} 
        \arrow[r, "T^{\mathcal{X}}", yshift=20pt]
        & \includegraphics[trim={4cm 3cm 4cm 6cm}, clip, width=0.25\columnwidth]{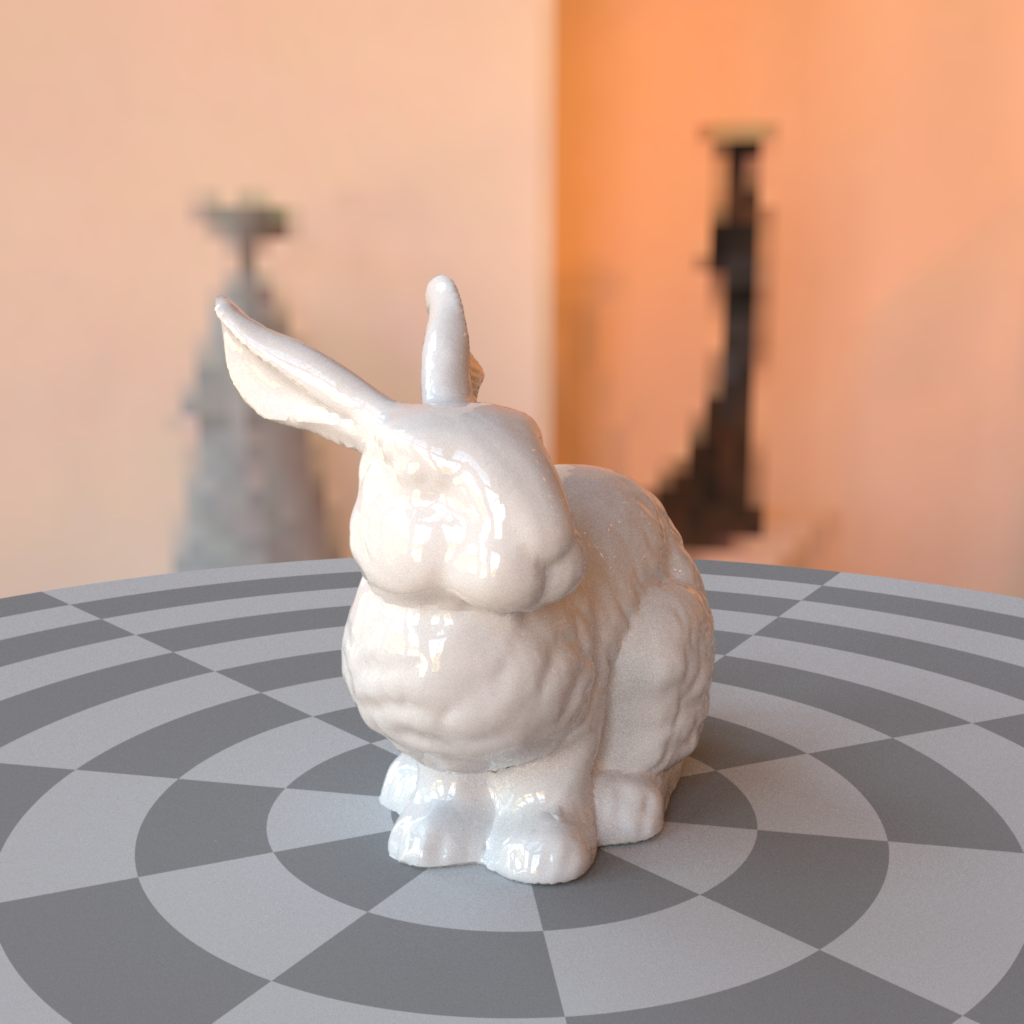} 
    \end{tikzcd}
    \vspace{-2pt}
    \caption{Rotating a mesh with $T^{\mathcal{Z}}$ and rendering it with $g$ is equivalent to rendering the original mesh and applying a transformation $T^{\mathcal{X}}$ in image space. This is true regardless of the choice of scene representation and rendering function.}
    \label{equivariance-diagram}
\end{center}
\vspace{-10pt}
\end{figure}

Indeed, we can consider applying some transformation $T^{\mathcal{Z}}$ to a scene representation. For example, we can rotate and translate a 3D mesh. The resulting image rendered by $g$ should then reflect these transformations, that is we would expect an equivalent transformation $T^{\mathcal{X}}$ to occur in image space (see Fig. \ref{equivariance-diagram}). We can write down this relation as%Similarly, if we apply a transformation $T^{\mathcal{X}}$ to an image, the scene representation inferred by the inverse renderer $f$ should undergo an equivalent transformation  $T^{\mathcal{Z}}$. We can write down these relations as

\begin{equation}
\label{equivariance-equations}
\begin{aligned}
% T^{\mathcal{Z}} f(\mathbf{x}) &= f(T^{\mathcal{X}} \mathbf{x})\\
T^{\mathcal{X}} g(\mathbf{z}) &= g(T^{\mathcal{Z}} \mathbf{z}).  
\end{aligned}
\end{equation}

This equation encodes the fact that transforming a scene representation with $T^{\mathcal{Z}}$ and rendering it with $g$ is equivalent to rendering the original scene and performing a transformation $T^{\mathcal{X}}$ on the rendered image. More specifically, the renderer is equivariant with respect to the transformations in image and scene space\footnote{Formally, $T^{\mathcal{X}}$ and $T^{\mathcal{Z}}$ represent the action of a group, such as the group of 3D rotations SO(3) or the group of 3D rotations and translations SE(3).}. We then define an \textit{equivariant scene representation} as one that satisfies the equivariance relation in equation (\ref{equivariance-equations}). We can therefore think of equivariant scene representations as a generalization of several other scene representations. Indeed, meshes, voxels, point clouds (and so on) paired with their appropriate rendering function all satisfy this equation.

%This equation encodes the fact that the renderer is equivariant with respect to the transformations in image and scene space\footnote{Formally, $T^{\mathcal{X}}$ and $T^{\mathcal{Z}}$ are required to be representations of a group, such as the group of 3D rotations SO(3) or the group of 3D rotations and translations SE(3).}. We then define an \textit{equivariant scene representation} as one that satisfies the equivariance relation in equation (\ref{equivariance-equations}). We can therefore think of equivariant scene representations as a generalization of several other scene representations. Indeed, meshes, voxels, point clouds (and so on) paired with their appropriate rendering function all satisfy this equation.

\section{Model} \label{model-section}

% Maybe this section should be more like how do we learn such a model? E.g. what data should you choose, what loss should you choose?
In this section, we design a model and loss that can be used to learn equivariant scene representations from data. While our formulation applies to general transformations and scene representations, we focus on the case where the scene representations are deep voxels and the family of transformations is 3D rotations. Specifically, we set $\mathcal{Z}=\mathbb{R}^{c_s \times d_s \times h_s \times w_s}$ where $c_s$, $d_s$, $h_s$, $w_s$ are the channels, depth, height and width of the scene representation. We denote the rotation operation in scene space by $\rotscene$ and the equivalent rotation operation acting on rendered images $\mathbf{x}$ by $\rotrender$.

As our model learns implicit scene representations, we do not require 3D ground truths. Instead, our dataset is composed of pairs of views of scenes and relative camera transformations linking the two views. Specifically, we assume the camera observing the scenes is on a sphere looking at the origin. For a given scene, we consider two image captures of the scene $\mathbf{x}_1$ and $\mathbf{x}_2$ and the \textit{relative} camera transformation between the two $\boldsymbol{\theta}=\theta \hat{\mathbf{n}}$ where $\theta$ is the angle and $\hat{\mathbf{n}}$ the axis parameterizing the 3D rotation\footnote{We use the axis-angle parameterization for notational convenience, but any rotation formalism such as euler angles, rotation matrices and quaternions could be used. In our implementation, we parameterize this rotation by a rotation matrix.}. A training data point is then given by $(\mathbf{x}_1, \mathbf{x}_2, \boldsymbol{\theta})$. In practice, we capture a large number of views for each scene and randomly sample new pairs at every iteration in training. This allows us to build models that generalize well across a large variety of camera transformations. 

\begin{figure}[t]
\begin{center}
\centerline{\includegraphics[width=\columnwidth]{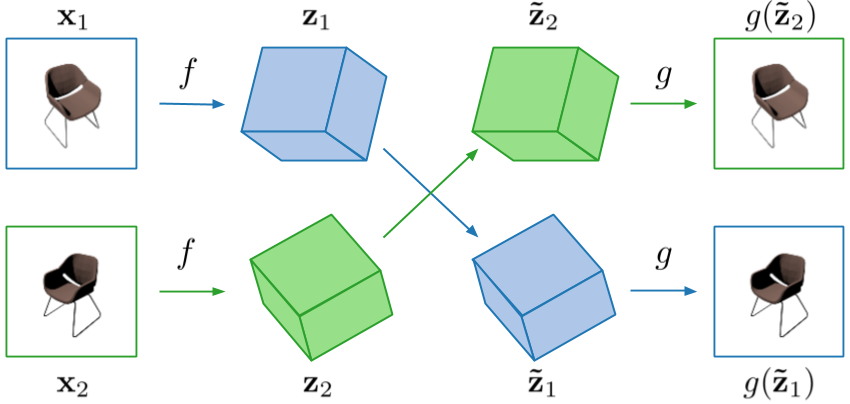}}
%\vspace{-5pt}
\caption{Model training. We encode two images $\mathbf{x}_1$, $\mathbf{x}_2$ of the \textit{same scene} into their respective scene representations $\mathbf{z}_1$, $\mathbf{z}_2$. Since they are representations of the same scene viewed from different points, we can rotate each one into the other. The rotated scene representations $\mathbf{\tilde{z}}_1$, $\mathbf{\tilde{z}}_2$ should then be decoded to match the swapped image pairs $\mathbf{x}_2$, $\mathbf{x}_1$.}
\label{model-training}
\end{center}
\vspace{-5pt}
\end{figure}

To design a loss that enforces equivariance with respect to the rotation transformation, we consider two images of the \textit{same scene} and their relative transformation $(\mathbf{x}_1, \mathbf{x}_2, \boldsymbol{\theta})$. We first map the images through the inverse renderer to obtain their scene representations $\mathbf{z}_1 = f(\mathbf{x}_1)$ and $\mathbf{z}_2 = f(\mathbf{x}_2)$. We then rotate each encoded representation by its relative transformation $\rotdeltascene$, such that $\mathbf{\tilde{z}}_1 = \rotdeltascene \mathbf{z}_1$ and $\mathbf{\tilde{z}}_2 = (\rotdeltascene)^{-1} \mathbf{z}_2$. As $\mathbf{z}_1$ and $\mathbf{z}_2$ represent the same scene in different poses, we expect the rotated $\mathbf{\tilde{z}}_1$ to be rendered as the image $\mathbf{x}_2$ and the rotated $\mathbf{\tilde{z}}_2$ as $\mathbf{x}_1$. This is illustrated in Fig. \ref{model-training}. We can then ensure our model obeys these transformations by minimizing

\begin{equation} \label{render-loss}
\mathcal{L}_\text{render} = || \mathbf{x}_2 - g(\mathbf{\tilde{z}}_1) || + || \mathbf{x}_1 - g(\mathbf{\tilde{z}}_2) ||.
\end{equation}

As $\mathbf{x}_2 = \rotdeltarender \mathbf{x}_1$, minimizing this loss then corresponds to satisfying the equivariance property for the renderer $g$. Note that the form of this loss function is similar to the ones proposed by \citeauthor{worrall2017interpretable} and \citeauthor{olszewski2019transformable}. % While this loss enforces equivariance of $g$, it does not necessarily enforce equivariance of the inverse renderer $f$. We can therefore define an analogous loss that enforces equivariance of the inverse renderer with respect to rotations

%\begin{equation} \label{scene-loss}
%\mathcal{L}_\text{scene} = || f(\mathbf{x}_2) - \mathbf{\tilde{z}}_1 ||_2 + || f(\mathbf{x}_1) - \mathbf{\tilde{z}}_2 ||_2.
%\end{equation}

%In practice we find that minimizing the rendering loss in equation (\ref{render-loss}) also minimizes the scene loss in equation (\ref{scene-loss}), making it unnecessary to explicitly minimize (\ref{scene-loss}). We instead use the scene loss as a metric to track training progress.

%The total loss is then a weighted sum $\mathcal{L} = \mathcal{L}_\text{render} + \lambda \mathcal{L}_\text{scene}$ ensuring that both the inverse and forward renderer are equivariant with respect to rotations of the camera.

\begin{figure}[t]
\begin{center}
\centerline{\includegraphics[width=\columnwidth]{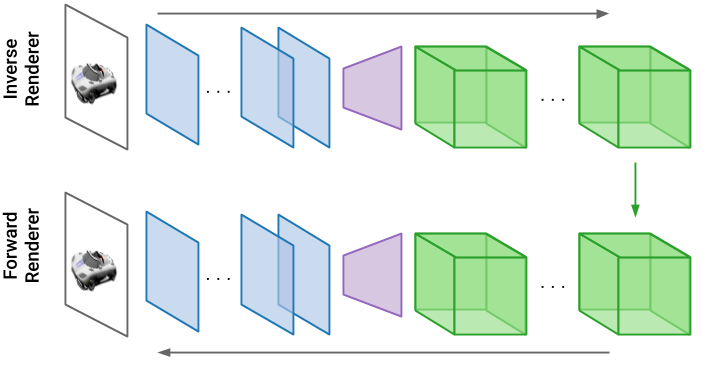}}
%\vspace{-5pt}
\caption{Model architecture. An input image (top left) is mapped through 2D convolutions (blue), followed by an inverse projection (purple) and a set of 3D convolutions (green). The inferred scene is then rendered through the transpose of this architecture.}
\label{model-architecture}
\end{center}
\vspace{-5pt}
\end{figure}

\textbf{Model architecture.} In contrast to most other works learning implicit scene representations \cite{worrall2017interpretable, Eslami1204, Chen_2019_ICCV}, our representation is spatial in three dimensions, allowing us to use fully convolutional architectures for both the inverse and forward neural renderer. To build the forward renderer, we take inspiration from RenderNet \cite{nguyen2018rendernet} and HoloGAN \cite{nguyen2019hologan} as these have been shown to achieve good performance on rendering tasks. Specifically, the scene representation $\mathbf{z}$ is mapped through a set of 3D convolutions, followed by a projection layer of $1 \times 1$ convolutions and finally a set of 2D convolutions mapping the projection to an image. The inverse renderer is simply defined as the transpose of this architecture (see Fig. \ref{model-architecture}). For complete details of the architecture, please refer to the appendix.

\textbf{Voxel rotation.} Defining the rotation operation in scene space $\rotscene$ is crucial. As our scene representation $\mathbf{z}$ is a deep voxel grid, we simply apply a 3D rotation matrix to the coordinates of the features in the voxel grid. As the rotated points may not align with the grid, we use inverse warping with trilinear interpolation to reconstruct the values at the voxel locations (see \citeauthor{szeliski2010computer} for more detail). We note that warping and interpolation operations are available in frameworks such as Pytorch and Tensorflow, making it simple to implement voxel rotations in practice.
% Specifically, we use inverse warping by applying the inverse rotation to voxels in the target volume to determine which values in the input volume will be used for interpolation. 

\textbf{Rendering loss.} There are several possible choices for the rendering loss, the most common being the $\ell_1$ norm, $\ell_2$ norm and SSIM \cite{wang2004image} or combinations thereof. As noted in other works \cite{worrall2017interpretable, snell2017learning} a weighted sum of $\ell_1$ and SSIM works well in practice. However, we found that our model is not particularly sensitive to the choice of regression loss, and analyse the various trade offs through ablation studies in the experimental section.

\section{Experiments}

We perform experiments on ShapeNet benchmarks \cite{chang2015shapenet} as well as on two new datasets designed to challenge the model on more complex scenes. For all experiments, the images are of size $128 \times 128$ and the scene representations are of size $64 \times 32 \times 32 \times 32$. For both the 2D and 3D parts of the network we use residual layers for convolutions that preserve the dimension of the input and strided convolutions for downsampling layers. We use the LeakyReLU nonlinearity \cite{maas2013rectifier} and GroupNorm \cite{wu2018group} for normalization. Complete architecture and training details can be found in the appendix.

Most novel view synthesis works are tested on the ShapeNet dataset or variants of it. However, renders from ShapeNet objects are typically very far from real life scenes, which tends to limit the use cases for models trained on them. As our scene representation and rendering framework make no restricting assumptions about the rendering process (such as requiring single objects, no reflections, no background etc.), we create new datasets to test the performance of our model on more advanced tasks.

The new datasets are challenging by design and are composed of photorealistic 3D scenes and 3D landscapes with textures from satellite images. We achieve compelling results on these datasets and hope they will spur further research into scene representations that are not limited to simple scenes without backgrounds. The code and datasets are available at \url{https://github.com/apple/ml-equivariant-neural-rendering}.

\begin{table}[t]
\begin{center}
\begin{small}
\begin{sc}
\begin{tabular}{l|cccc}
\toprule
  & TCO & dGQN & SRN & Ours \\
\midrule
\begin{tabular}{@{}l@{}}Requires absolute \\ pose\end{tabular} & \textcolor{red}{Yes} & \textcolor{red}{Yes} & \textcolor{red}{Yes} & \textcolor{darkgreen}{No} \\
\hline
\begin{tabular}{@{}l@{}}Requires pose at \\ inference time\end{tabular} & \textcolor{darkgreen}{No} & \textcolor{red}{Yes} & \textcolor{red}{Yes} & \textcolor{darkgreen}{No} \\
\hline
\begin{tabular}{@{}l@{}}Optimization at \\ inference time\end{tabular} & \textcolor{darkgreen}{No} & \textcolor{darkgreen}{No} & \textcolor{red}{Yes} & \textcolor{darkgreen}{No} \\
\bottomrule
\end{tabular}
\end{sc}
\end{small}
\caption{Requirements for each baseline. Our model performs comparably to other models that make much stronger assumptions about the data and inference process.}
\label{model-assumptions}
\end{center}
\vspace{-10pt}
\end{table}

\subsection{Baselines} \label{baseline-section}

We compare our model with three strong baselines. The first is the model proposed by \citeauthor{tatarchenko2016multi} which we refer to as TCO, the second is a deterministic variant of Generative Query Networks \cite{Eslami1204} which we refer to as dGQN and the third is the Scene Representation Network (SRN) as described in \citeauthor{sitzmann2019scene}\footnote{For detailed descriptions of these baselines, please refer to the appendix of \cite{sitzmann2019scene}.}. All baselines make strong assumptions that substantially simplify the view-synthesis and scene representation problem. We discuss each of these assumptions in detail below and provide a comparison in Table \ref{model-assumptions}. Our model requires neither of these assumptions, making the task it has to solve considerably more challenging while also being more generally applicable.

\textbf{Absolute and relative pose.} All baselines require an absolute coordinate system\footnote{This is often referred to as a world coordinate system.} for the pose (or viewpoints). For example, when trained on chairs, the viewpoint corresponding to the camera being at the origin would be the one observing the chair face on. The poses are then absolute in the sense that the camera at the origin corresponds to observing the chair face on for all chairs, i.e. we need all scenes to be perfectly aligned. While this is possible for simple datasets like ShapeNet, it is difficult to define a consistent alignment for a set of scenes, particularly for complex scenes with backgrounds and real life images. In contrast, our model does not require any notion of alignment or absolute pose. Equivariance is exactly why we are able to build a representation that is “origin-free”, because it only depends on relative transformations between poses.

\textbf{Pose at inference time.} In order to infer a scene representation, our model takes as input a single image of the scene. In contrast, both dGQN and SRNs require an image as well as the viewpoint from which the image was taken. This considerably simplifies the task as the model does not need to infer the pose.

\textbf{Optimization at inference time.} At inference time, SRNs require solving an optimization problem in order to fit a scene to the model. As such, inferring a scene representation from a single input image (on a Tesla V100 GPU) takes 2 minutes with SRNs but only 22ms for our model (three orders of magnitude faster). The idea of training at inference time is a crucial element of SRNs and other works in 3D computer vision \cite{DeepSDF}, but is not required for our model.

\subsection{Chairs}

We evaluate our model on the ShapeNet chairs class by following the experimental setup given in \citeauthor{sitzmann2019scene}, using the same train/validation/test splits. The dataset is composed of 6591 chairs each with 50 views sampled uniformly on the sphere for a total of 329,550 images. Images are sampled on the \textit{full sphere} around the object, making the task much more difficult than typical setups which limit the elevation or azimuth or both \cite{tatarchenko2016multi, Chen_2019_ICCV, olszewski2019transformable}.

\begin{figure}[t]
\begin{center}

\centerline{
    \raisebox{1.25pt}{\includegraphics[width=0.159\columnwidth]{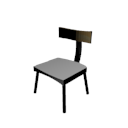}}%
    \hspace{3pt}%
    \textcolor{darkgray1}{\vrule width 1pt height 35pt depth -4pt}%
    \hspace{2pt}%
    \includegraphics[width=0.83\columnwidth]{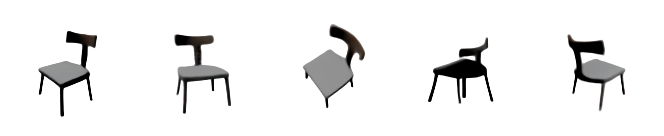}
}

%\centerline{
%    \raisebox{1.25pt}{\includegraphics[width=0.159\columnwidth]{figures/chair-input-2.png}}%
%    \hspace{3pt}%
%    \textcolor{darkgray1}{\vrule width 1pt height 35pt depth -4pt}%
%    \hspace{2pt}%
%    \includegraphics[width=0.83\columnwidth]{figures/chair-novel-2.png}
%}

\centerline{
    \raisebox{1.25pt}{\includegraphics[width=0.159\columnwidth]{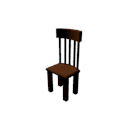}}%
    \hspace{3pt}%
    \textcolor{darkgray1}{\vrule width 1pt height 35pt depth -4pt}%
    \hspace{2pt}%
    \includegraphics[width=0.83\columnwidth]{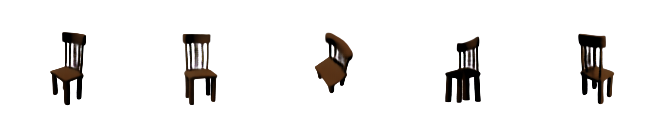}
}

\centerline{
    \raisebox{1.25pt}{\includegraphics[width=0.159\columnwidth]{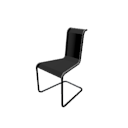}}%
    \hspace{3pt}%
    \textcolor{darkgray1}{\vrule width 1pt height 35pt depth -4pt}%
    \hspace{2pt}%
    \includegraphics[width=0.83\columnwidth]{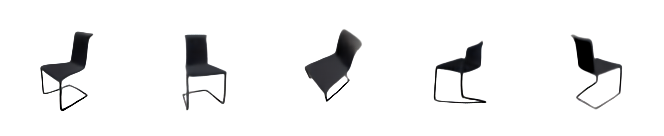}
}

\caption{Novel view synthesis for chairs. Given a single image of an unseen object (left), we infer a scene representation, rotate and render it with our learned renderer to generate novel views. Due to space constraints we include chairs with interesting properties here and show randomly sampled chairs in the appendix.}
\label{novel-view-chairs}
\end{center}
\vspace{-15pt}
\end{figure}

\textbf{Novel view synthesis.} Results for novel view synthesis are shown in Fig. \ref{novel-view-chairs}. The novel views were produced by taking a single image of an \textit{unseen chair}, inferring its scene representation with the inverse renderer, rotating the scene and generating a novel view with the learned neural renderer. As can be seen, our model is able to generate plausible views of new chairs even when viewed from difficult angles and in the presence of occlusion. The model works well even for oddly shaped chairs with thin structures.

% Novel view synthesis

\textbf{Quantitative comparisons.} To perform quantitative comparisons, we follow the setup in \citeauthor{sitzmann2019scene} by considering a single informative view of an unseen test object and measuring the reconstruction performance on the upper hemisphere around the object (results are shown in Table \ref{psnr-table}). Surprisingly, even though our model makes much weaker assumptions than all the baselines, it significantly improves upon both the TCO and dGQN baselines and is comparable with the state of the art SRNs. %(22.83dB vs 22.89dB).

% Quantitative
\begin{table}[t]
\begin{center}
\begin{small}
\begin{sc}
\begin{tabular}{lcccc}
\toprule
Dataset & TCO & dGQN & SRN & Ours \\
\midrule
Chairs & 21.27 & 21.59 & 22.89 & 22.83 \\
\bottomrule
\end{tabular}
\end{sc}
\end{small}
\caption{Reconstruction accuracy (higher is better) in PSNR (units of dB) for baselines and our model on ShapeNet chairs.}
\label{psnr-table}
\end{center}
%\vspace{-15pt}
\end{table}

\begin{figure}[t]
\small \hspace{5pt} Input \qquad \hspace{2pt} dGQN \qquad \hspace{-5pt} TCO \qquad \hspace{-2pt} SRN \qquad \hspace{-1pt} Ours \qquad \hspace{-1pt} Target
\vspace{-4pt}
\begin{center}
\centerline{
    \raisebox{1.25pt}{\includegraphics[width=0.159\columnwidth]{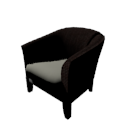}}%
    \hspace{3pt}%
    \textcolor{darkgray1}{\vrule width 1pt height 35pt depth -4pt}%
    \hspace{2pt}%
    \includegraphics[width=0.83\columnwidth]{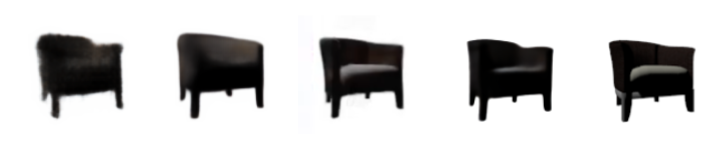}
}

\centerline{
    \raisebox{1.25pt}{\includegraphics[width=0.159\columnwidth]{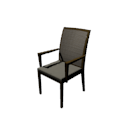}}%
    \hspace{3pt}%
    \textcolor{darkgray1}{\vrule width 1pt height 35pt depth -4pt}%
    \hspace{2pt}%
    \includegraphics[width=0.83\columnwidth]{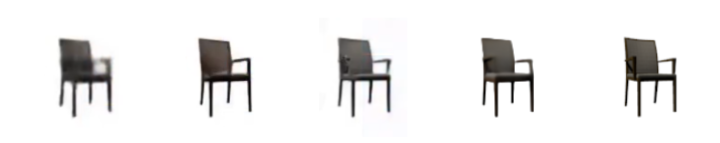}
}

\centerline{
    \raisebox{1.25pt}{\includegraphics[width=0.159\columnwidth]{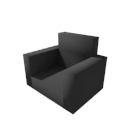}}%
    \hspace{3pt}%
    \textcolor{darkgray1}{\vrule width 1pt height 35pt depth -4pt}%
    \hspace{2pt}%
    \includegraphics[width=0.83\columnwidth]{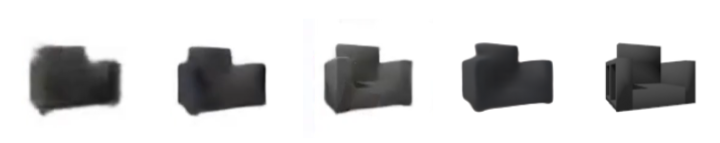}
}

\caption{Qualitative comparisons for single shot novel view synthesis. The baseline images were borrowed with permission from \citeauthor{sitzmann2019scene}.}
\label{qualitative-comparisons}
\end{center}
\vspace{-15pt}
\end{figure}

\textbf{Qualitative comparisons.} We show qualitative comparisons with the baselines for single shot novel view synthesis in Fig. \ref{qualitative-comparisons}. As can be seen our model produces high quality novel views that are comparable to or better than dGQN and TCO while being slightly worse than SRNs.

% Qualitative

\subsection{Cars}

We also evaluate our model on the ShapeNet cars class, allowing us to test our model on images with richer texture than chairs. The dataset is composed of 3514 cars each with 50 views sampled uniformly on the sphere for a total of 175,700 images.

\begin{figure}[t]
\begin{center}

\centerline{
    \raisebox{1.25pt}{\includegraphics[width=0.159\columnwidth]{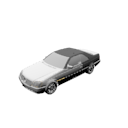}}%
    \hspace{3pt}%
    \textcolor{darkgray1}{\vrule width 1pt height 35pt depth -4pt}%
    \hspace{2pt}%
    \includegraphics[width=0.83\columnwidth]{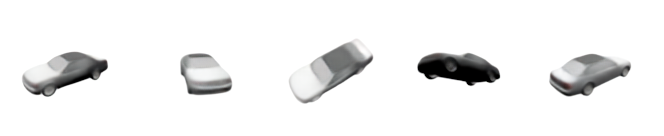}
}

\centerline{
    \raisebox{1.25pt}{\includegraphics[width=0.159\columnwidth]{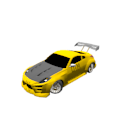}}%
    \hspace{3pt}%
    \textcolor{darkgray1}{\vrule width 1pt height 35pt depth -4pt}%
    \hspace{2pt}%
    \includegraphics[width=0.83\columnwidth]{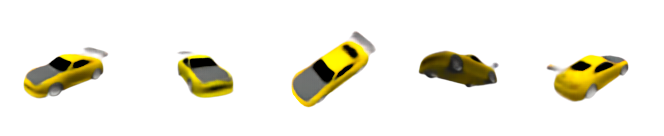}
}

\centerline{
    \raisebox{1.25pt}{\includegraphics[width=0.159\columnwidth]{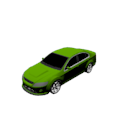}}%
    \hspace{3pt}%
    \textcolor{darkgray1}{\vrule width 1pt height 35pt depth -4pt}%
    \hspace{2pt}%
    \includegraphics[width=0.83\columnwidth]{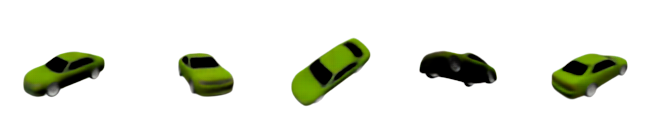}
}

\caption{Novel view synthesis for cars.}
\label{novel-view-cars}
\end{center}
%\vspace{-5pt}
\end{figure}

\textbf{Novel view synthesis}. As can be seen in Fig. \ref{novel-view-cars}, our model is able to generate plausible views for cars with various colors and thin structures like spoilers. While our model successfully infers 3D shape and appearance, it still struggles to capture some fine texture and geometry details (see Section \ref{scope-section} for a thorough discussion of the limitations and failures of our model).

% Novel view synthesis

\begin{figure}[t]
\small \hspace{24pt} Input \qquad \hspace{2pt} SRN \quad \hspace{-7pt} SRN (relative) \quad \hspace{-7pt} Ours \quad \hspace{2pt} Target
\vspace{-4pt}
\begin{center}
\centerline{
    \raisebox{1.25pt}{\includegraphics[width=0.159\columnwidth]{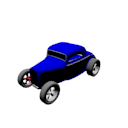}}%
    \hspace{3pt}%
    \textcolor{darkgray1}{\vrule width 1pt height 35pt depth -4pt}%
    \hspace{2pt}%
    \raisebox{1.25pt}{\includegraphics[width=0.159\columnwidth]{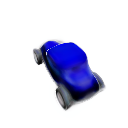}}
    \raisebox{1.25pt}{\includegraphics[width=0.159\columnwidth]{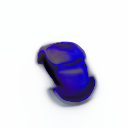}}
    \raisebox{1.25pt}{\includegraphics[width=0.159\columnwidth]{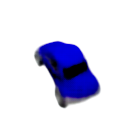}}
    \raisebox{1.25pt}{\includegraphics[width=0.159\columnwidth]{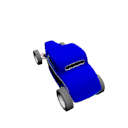}}
}

\centerline{
    \raisebox{1.25pt}{\includegraphics[width=0.159\columnwidth]{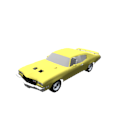}}%
    \hspace{3pt}%
    \textcolor{darkgray1}{\vrule width 1pt height 35pt depth -4pt}%
    \hspace{2pt}%
    \raisebox{1.25pt}{\includegraphics[width=0.159\columnwidth]{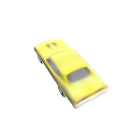}}
    \raisebox{1.25pt}{\includegraphics[width=0.159\columnwidth]{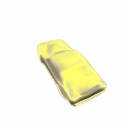}}
    \raisebox{1.25pt}{\includegraphics[width=0.159\columnwidth]{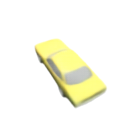}}
    \raisebox{1.25pt}{\includegraphics[width=0.159\columnwidth]{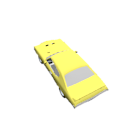}}
}

\centerline{
    \raisebox{1.25pt}{\includegraphics[width=0.159\columnwidth]{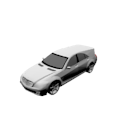}}%
    \hspace{3pt}%
    \textcolor{darkgray1}{\vrule width 1pt height 35pt depth -4pt}%
    \hspace{2pt}%
    \raisebox{1.25pt}{\includegraphics[width=0.159\columnwidth]{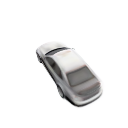}}
    \raisebox{1.25pt}{\includegraphics[width=0.159\columnwidth]{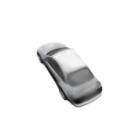}}
    \raisebox{1.25pt}{\includegraphics[width=0.159\columnwidth]{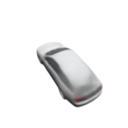}}
    \raisebox{1.25pt}{\includegraphics[width=0.159\columnwidth]{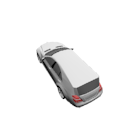}}
}

\caption{Qualitative comparisons on cars between SRNs, SRNs with relative poses around the up axis and our model.}
\label{qualitative-comparisons-cars}
\end{center}
\vspace{-15pt}
\end{figure}

\textbf{Absolute and relative poses.} As mentioned in Section \ref{baseline-section}, our model only relies on relative transformations and therefore alleviates the need for alignment between scenes. As all baselines require absolute poses and alignment between scenes, we run tests to see how important this assumption is. Specifically, we break the alignment between scenes in the cars dataset by randomly rotating each scene around the up axis\footnote{We found that rotating around one axis was enough to see a significant effect. Rotating around all 3 axes would likely have an even larger effect.}. We then train an SRN model on the perturbed and unperturbed dataset to understand to which extent the model relies on the absolute coordinates. As can be seen in Fig. \ref{qualitative-comparisons-cars}, breaking the alignment between scenes significantly deteriorates the performance of SRNs while it leaves the performance of our model unchanged. This is similarly reflected when measuring reconstruction accuracy on the test set (see Table \ref{psnr-table-cars}).

% Qualitative

\begin{table}[t]
\begin{center}
\begin{small}
\begin{sc}
\begin{tabular}{lccc}
\toprule
Dataset & SRN & SRN (relative) & Ours \\
\midrule
Cars & 22.36 & 21.05 & 22.26 \\
\bottomrule
\end{tabular}
\end{sc}
\end{small}
\caption{Reconstruction accuracy (higher is better) in PSNR (units of dB) on ShapeNet cars.}
\label{psnr-table-cars}
\end{center}
\end{table}

\subsection{MugsHQ}

As the model does not make any restricting assumptions about the rendering process, we test it on more difficult scenes by building the MugsHQ dataset based on the mugs class from ShapeNet. Instead of rendering images on a blank background, every scene is rendered with an environment map (lighting conditions) and a checkerboard disk platform. For each of the 214 mugs, we sample 150 viewpoints uniformly over the upper hemisphere and render views using the Mitsuba renderer \cite{Mitsuba}. Note that the environment map and disk platform is the same for every mug. The resulting scenes include more complex visual effects like reflections and look more realistic than typical ShapeNet renders, making the task of novel view synthesis considerably more challenging. A complete description of the dataset as well as samples can be found in the appendix.

\begin{figure}[t]
\begin{center}

\centerline{
    \raisebox{1.25pt}{\includegraphics[width=0.159\columnwidth]{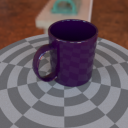}}%
    \hspace{3pt}%
    \textcolor{darkgray1}{\vrule width 1pt height 35pt depth -4pt}%
    \hspace{2pt}%
    \includegraphics[width=0.83\columnwidth]{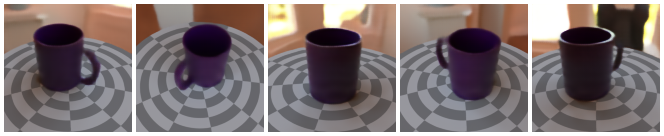}
}

\centerline{
    \raisebox{1.25pt}{\includegraphics[width=0.159\columnwidth]{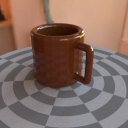}}%
    \hspace{3pt}%
    \textcolor{darkgray1}{\vrule width 1pt height 35pt depth -4pt}%
    \hspace{2pt}%
    \includegraphics[width=0.83\columnwidth]{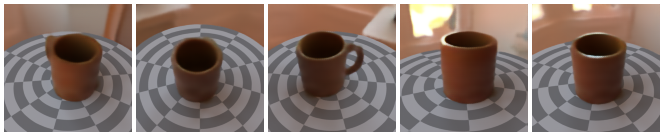}
}

\centerline{
    \raisebox{1.25pt}{\includegraphics[width=0.159\columnwidth]{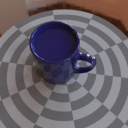}}%
    \hspace{3pt}%
    \textcolor{darkgray1}{\vrule width 1pt height 35pt depth -4pt}%
    \hspace{2pt}%
    \includegraphics[width=0.83\columnwidth]{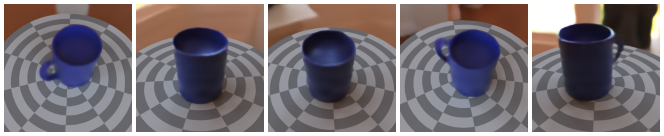}
}

\vspace{-5pt}
\caption{Novel view synthesis on MugsHQ.}
\label{novel-view-mugs-hq}
\end{center}
\vspace{-10pt}
\end{figure}

\textbf{Novel view synthesis.} Results for single shot novel view synthesis on unseen mugs are shown in Fig. \ref{novel-view-mugs-hq}. As can be seen, the model successfully infers the shape of unseen mugs from a single image and is able to perform large viewpoint transformations. Even from difficult viewpoints, the model is able to produce consistent and realistic views of the scenes, even generating reflections on the mug edges. As is the case for the ShapeNet dataset, our model can still miss fine details such as thin mug handles and struggles with some oddly shaped mugs (see Section \ref{scope-section} for examples).

% Novel view synthesis

\subsection{Mountains}

% Describe dataset and why it's really challenging. This is beyond the scope of any algorithm we are familiar with and hope that this dataset can be useful for pushing the boundaries of research in neural rendering. We present results for our model and hope that this can serve as a useful baseline for future comparisons.

We also introduce 3D mountains, a dataset of mountain landscapes. We created the dataset by scraping the height, latitude and longitude of the 559 highest mountains in the Alps (we chose this mountain range because it was easiest to find data). We then used satellite images combined with topography data to sample random views of each mountain at a fixed height (see appendix for samples and detailed description). This dataset is extremely challenging, with varied and complex geometry and texture. While obtaining high quality results on this dataset is beyond the scope of our algorithm, we hope it can be useful for pushing the boundaries of research in neural rendering.

\begin{figure}[t]
\begin{center}
\centerline{
    \raisebox{1.25pt}{\includegraphics[width=0.159\columnwidth]{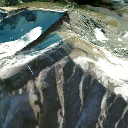}}%
    \hspace{3pt}%
    \textcolor{darkgray1}{\vrule width 1pt height 35pt depth -4pt}%
    \hspace{2pt}%
    \includegraphics[width=0.83\columnwidth]{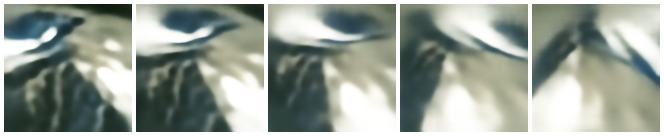}
}

\centerline{
    \raisebox{1.25pt}{\includegraphics[width=0.159\columnwidth]{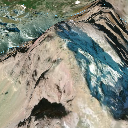}}%
    \hspace{3pt}%
    \textcolor{darkgray1}{\vrule width 1pt height 35pt depth -4pt}%
    \hspace{2pt}%
    \includegraphics[width=0.83\columnwidth]{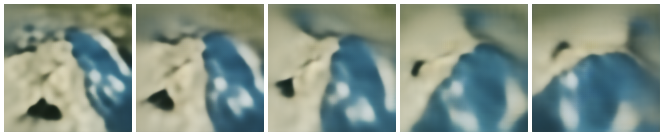}
}

\centerline{
    \raisebox{1.25pt}{\includegraphics[width=0.159\columnwidth]{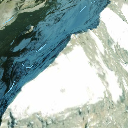}}%
    \hspace{3pt}%
    \textcolor{darkgray1}{\vrule width 1pt height 35pt depth -4pt}%
    \hspace{2pt}%
    \includegraphics[width=0.83\columnwidth]{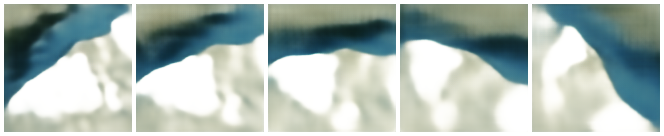}
}
\vspace{-5pt}
\caption{Novel view synthesis on 3D mountains.}
\label{novel-view-mountains}
\end{center}
\vspace{-5pt}
\end{figure}

\textbf{Novel view synthesis.} Results for single shot novel view synthesis are shown in Fig. \ref{novel-view-mountains}. While the model struggles to capture high frequency detail, it faithfully reproduces the 3D structure and texture of the mountain as the camera rotates around the scene representation. For a variety of mountain landscapes (snowy, rocky etc.), our model is able to generate plausible, albeit blurry, views. An interesting feature is that, for views near the input image, the generated images are considerably sharper than for views far away from the input. This is likely due to the considerable uncertainty in generating views far from the source view: given the front of a mountain, there are many plausible ways the back of the mountain could appear. As our model is deterministic, it generates sharper views near the input where there is less uncertainty and blurs views far from the input where there is more uncertainty. %We hope these results can serve as a useful baseline for future comparisons on this challenging dataset.

\begin{figure}[t]
\small $\ell_1$+SSIM \quad \hspace{2pt} $\ell_2$ \qquad \hspace{2pt} Target \qquad $\ell_1$+SSIM \quad \hspace{2pt} $\ell_2$ \qquad \hspace{2pt} Target \vspace{-4pt}
\begin{center}
\centerline{
    \includegraphics[width=0.159\columnwidth]{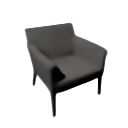}
    \includegraphics[width=0.159\columnwidth]{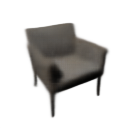}
    \includegraphics[width=0.159\columnwidth]{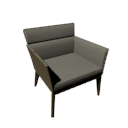}
    \hspace{5pt}
    \includegraphics[width=0.159\columnwidth]{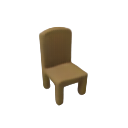}
    \includegraphics[width=0.159\columnwidth]{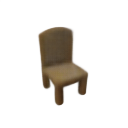}
    \includegraphics[width=0.159\columnwidth]{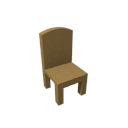}
}

\centerline{
    \includegraphics[width=0.159\columnwidth]{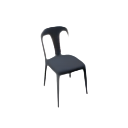}
    \includegraphics[width=0.159\columnwidth]{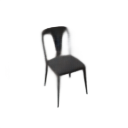}
    \includegraphics[width=0.159\columnwidth]{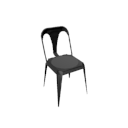}
    \hspace{5pt}
    \includegraphics[width=0.159\columnwidth]{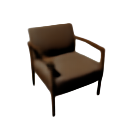}
    \includegraphics[width=0.159\columnwidth]{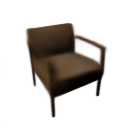}
    \includegraphics[width=0.159\columnwidth]{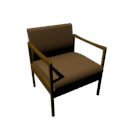}
}
\caption{Comparisons on chairs showing the trade off between different rendering losses.}
\label{rendering-loss-ablation}
\end{center}
\vspace{-10pt}
\end{figure}

\subsection{Ablation studies}

We perform ablation studies to test the trade offs between various rendering losses. Fig. \ref{rendering-loss-ablation} shows the difference in generated images when using $\ell_2$ and $\ell_1 + \text{SSIM}$ losses. While both losses perform well, the $\ell_2$ loss produces somewhat blurrier images than the $\ell_1 + \text{SSIM}$ loss. However, there are also cases where the $\ell_1 + \text{SSIM}$ produces artifacts that the $\ell_2$ loss does not. Ultimately, there is a trade off between using the two losses and the choice is largely dependent on the application.

% While the simple model formulation (training the model with an $\ell_2$ loss between the generated and ground truth images) described in Section \ref{model-section} works surprisingly well, we find that some additional changes provide small but consistent gains.

% \textbf{Rendering loss}. Fig. \ref{rendering-loss-ablation} shows the difference in generated images when using $\ell_2$ and $\ell_1 + \text{SSIM}$ losses. As can be seen, the $\ell_2$ loss produces somewhat blurrier images than the $\ell_1 + \text{SSIM}$ loss. However, there are also cases where the $\ell_1 + \text{SSIM}$ produces artifacts that the $\ell_2$ loss does not. Ultimately, there is a trade off between using the two losses and the choice is largely dependent on the application.

% \textbf{Equivariance of inverse renderer}. While training the model with the rendering loss in equation (\ref{render-loss}) enforces equivariance of the neural renderer it does not necessarily encourage equivariance of the inverse renderer. This motivated us to introduce the scene loss in equation (\ref{scene-loss}). As can be shown in Fig. \ref{scene-loss-ablation}, adding the scene loss, and thus encouraging the inverse renderer to be equivariant, decreases the rendering loss on validation data.

% We note also that this shows even the extremely simple model formulation we have works well. The scene loss and L1 + SSIM loss help, but even just training the model with an L2 rendering loss gives very convincing results.
% Insist on simple model working really well.
\vspace{-2pt}
\section{Scope, limitations and future work} \label{scope-section}

In this section, we discuss some of the advantages and weaknesses of our method as well as potential directions for future work.

\textbf{Advantages.} The main advantage of our model is that it makes very few assumptions about the scene representation and rendering process. Indeed, we learn representations simply by enforcing equivariance with respect to 3D rotations. As such, we can easily encode material, texture and lighting which is difficult with traditional 3D representations. The simplicity of our model also means that it can be trained purely from posed 2D images with no 3D supervision. As we have shown, this allows us to apply our method to interesting data where obtaining 3D geometry is difficult. Crucially, and unlike most other methods, our model does not require alignment between scenes nor any pose information at inference time. Further, our model is fast: inferring a scene representation simply corresponds to performing a forward pass of a neural network. This is in contrast to most other methods that require solving an expensive optimization problem at inference time for every new observed image \cite{nguyen2018rendernet, DeepSDF, sitzmann2019scene}. Rendering is also performed in a single forward pass, making it faster than other methods that often require recurrence to produce an image \cite{Eslami1204, sitzmann2019scene}.

\begin{figure}[t]
\small \hspace{3pt} Input \quad \hspace{5pt} Model \quad \hspace{2pt} Target \qquad \hspace{4pt} Input \quad \hspace{5pt} Model \quad \hspace{2pt} Target
\vspace{-4pt}
\begin{center}
\centerline{
    \includegraphics[width=0.159\columnwidth]{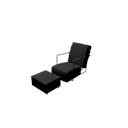}
    \includegraphics[width=0.159\columnwidth]{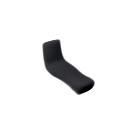}
    \includegraphics[width=0.159\columnwidth]{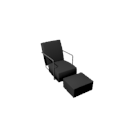}
    \hspace{5pt}
    \includegraphics[width=0.159\columnwidth]{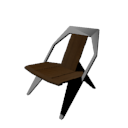}
    \includegraphics[width=0.159\columnwidth]{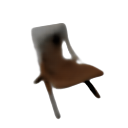}
    \includegraphics[width=0.159\columnwidth]{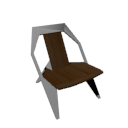}
}

\centerline{
    \includegraphics[width=0.159\columnwidth]{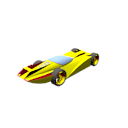}
    \includegraphics[width=0.159\columnwidth]{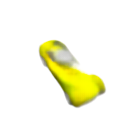}
    \includegraphics[width=0.159\columnwidth]{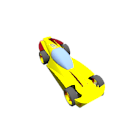}
    \hspace{5pt}
    \includegraphics[width=0.159\columnwidth]{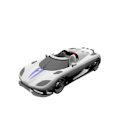}
    \includegraphics[width=0.159\columnwidth]{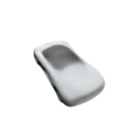}
    \includegraphics[width=0.159\columnwidth]{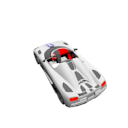}
}

\centerline{
    \includegraphics[width=0.159\columnwidth]{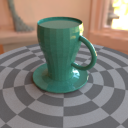}
    \includegraphics[width=0.159\columnwidth]{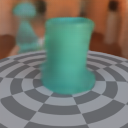}
    \includegraphics[width=0.159\columnwidth]{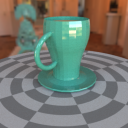}
    \hspace{5pt}
    \includegraphics[width=0.159\columnwidth]{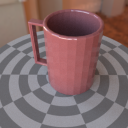}
    \includegraphics[width=0.159\columnwidth]{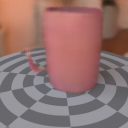}
    \includegraphics[width=0.159\columnwidth]{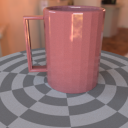}
}
\vspace{-5pt}
\caption{Failure examples of our model. As can be seen, the model fails on oddly shaped chairs, cars and mugs. On cars, the model sometimes infers the correct shape but misses high frequency texture detail. On mugs, the model can miss mug handles and other thin structure.}
\label{failure-examples}
\end{center}
\vspace{-20pt}
\end{figure}

\textbf{Limitations.} As our scene representation is spatial and 3-dimensional, our model is quite memory hungry. This implies we need to use a fairly small batch size which can make training slow (see appendix for detailed analysis of training times). Using a voxel-like representation could also make it difficult to generalize the model to other symmetries such as translations. In addition, our model typically produces samples of lower quality than models which make stronger assumptions. As an example, SRNs generally produce sharper and more detailed images than our model and are able to infer more fine-grained 3D information. Further SRNs can, unlike our model, generalise to viewpoints that were not observed during training (such as rolling the camera or zooming). While this is partly because we are solving a task that is inherently more difficult, it would still be desirable to narrow this gap in performance. We also show some failure cases of our model in Fig. \ref{failure-examples}. As can be seen, the model struggles with very thin structures as well as objects with unusual shapes. Further, the model can create unrealistic renderings in certain cases, such as mugs with disconnected handles.

\textbf{Future work.} The main idea of the paper is that equivariance with respect to symmetries of a real scene provides a strong inductive bias for representation learning of 3D environments. While we implement this using voxels as the representation and rotations as the symmetry, we could just as well have chosen point clouds as the representation and translation as the symmetry. The formulation of the model and loss are independent of the specific choices of representation and symmetry and we plan to explore the use of different representations and symmetries in future work. 

In addition, our model is deterministic, while inferring a scene from an image is an inherently uncertain process. Indeed, for a given image, there are several plausible scenes that could have generated it and, similarly, several different scenes could be rendered as the same image. It would therefore be interesting to learn a \textit{distribution} over scenes $p(\text{scene} | \text{image})$. Training a probabilistic or adversarial model may also help sharpen rendered images.

Another promising route would be to use the learned scene representation for 3D reconstruction. Indeed, most 3D reconstruction methods are object-centric (i.e. every object is reconstructed in the same orientation). This has been shown to cause models to effectively perform shape classification instead of reconstruction \cite{tatarchenko2019single}. As our scene representation is view-centric, it is likely that it could be useful for the downstream task of 3D reconstruction in the view-centric case.

\vspace{-2pt}
\section{Conclusion}

In this paper, we proposed learning scene representations by ensuring that they \textit{transform} like real 3D scenes. The proposed model requires no 3D supervision and can be trained using only posed 2D images. At test time, our model can, from a single image and in real time, infer a scene representation and manipulate this representation to render novel views. Finally, we introduced two challenging new datasets which we hope will help spur further research into neural rendering and scene representations for complex scenes. 

\section*{Acknowledgements}
We thank Shuangfei Zhai, Walter Talbott and Leon Gatys for useful discussions. We also thank Lilian Liang and Leon Gatys for help with running compute jobs. We thank Per Fahlberg for his help in generating the 3D mountains dataset. We are also grateful for Vincent Sitzmann for his help with generating ShapeNet datasets and benchmarks. We also thank Russ Webb for feedback on an early version of the manuscript. Finally we thank the anonymous reviewers for their useful feedback and suggestions.

\bibliography{icml_paper}
\bibliographystyle{icml2020}

%\iffalse

\newpage

\appendix

\section{Architecture details and hyperparameters}

The inverse renderer is composed of 3 submodels: a 2D convolutional network mapping images to 2D features, an inverse projection layer mapping 2D features to 3D features and a 3D convolutional network mapping 3D features to the scene representation. Each subnetwork is described in detail in the tables below. The renderer is simply the transpose of the inverse renderer with a sigmoid activation at the ouput layer to ensure pixel values are in $[0, 1]$.

Note that every layer is followed by a GroupNorm layer and a LeakyReLU activation (except the final scene and image layers). Each ResBlock is composed of a sequence of $1 \times 1$, $3 \times3$, $1 \times 1$ convolutions added to the identity.

To ensure rotations of the scene representation do not exit the bounds of the voxel grid we apply a spherical mask to the scene representation before performing rotations.

The full model has 11.5 million trainable parameters.

\begin{table}[h]
\begin{center}
\begin{small}
\begin{sc}
\begin{tabular}{ccc}
\toprule
Input shape & Output shape & Operation \\
\midrule
(3, 128, 128) & (64, 128, 128) & 1x1 Conv \\
(64, 128, 128) & (64, 128, 128) & 2x ResBlock \\
(64, 128, 128) & (128, 64, 64) & 4x4 Conv, Stride 2 \\
(128, 64, 64) & (128, 64, 64) & 1x ResBlock \\
(128, 64, 64) & (128, 32, 32) & 4x4 Conv, Stride 2 \\
(128, 32, 32) & (128, 32, 32) & 1x ResBlock \\
(128, 32, 32) & (256, 16, 16) & 4x4 Conv, Stride 2 \\
(256, 16, 16) & (256, 16, 16) & 1x ResBlock \\
(256, 16, 16) & (128, 32, 32) & 4x4 Conv.T, Stride 2 \\
(128, 32, 32) & (128, 32, 32) & 2x ResBlock \\
\bottomrule
\end{tabular}
\end{sc}
\end{small}
\end{center}
\vspace{-5pt}
\caption{Architecture of 2D subnetwork.}
\vspace{5pt}
\end{table}

\begin{table}[h]
\begin{center}
\begin{small}
\begin{sc}
\begin{tabular}{ccc}
\toprule
Input shape & Output shape & Operation \\
\midrule
(128, 32, 32) & (256, 32, 32) & 1x1 Conv \\
(256, 32, 32) & (512, 32, 32) & 1x1 Conv \\
(512, 32, 32) & (1024, 32, 32) & 1x1 Conv \\
(1024, 32, 32) & (32, 32, 32, 32) & Reshape \\
\bottomrule
\end{tabular}
\end{sc}
\end{small}
\end{center}
\vspace{-5pt}
\caption{Architecture of inverse projection network from 2D to 3D.}
\vspace{5pt}
\end{table}

\begin{table}[h]
\begin{center}
\begin{small}
\begin{sc}
{\setlength{\tabcolsep}{0.4em}
\begin{tabular}{ccc}
\toprule
Input shape & Output shape & Operation \\
\midrule
(32, 32, 32, 32) & (32, 32, 32, 32) & 1x1 Conv \\
(32, 32, 32, 32) & (32, 32, 32, 32) & 2x ResBlock \\
(32, 32, 32, 32) & (128, 16, 16, 16) & 4x4 Conv, Stride 2 \\
(128, 16, 16, 16) & (128, 16, 16, 16) & 2x ResBlock \\
(128, 16, 16, 16) & (64, 32, 32, 32) & 4x4 Conv.T, Stride 2 \\
(64, 32, 32, 32) & (64, 32, 32, 32) & 2x ResBlock \\
\bottomrule
\end{tabular}}
\end{sc}
\end{small}
\end{center}
\vspace{-5pt}
\caption{Architecture of 3D subnetwork.}
\vspace{5pt}
\end{table}

\textbf{Hyperparameters.} When training with $\ell_1 + \text{SSIM}$ loss, we set the weight of the SSIM loss to 0.05.

\textbf{Training.} We train each model for 100 epochs on all datasets, although most models converge much earlier than this (around 60 epochs). When training on a single GPU we use a batch size of 16 and when training on 8 GPUs we use a batch size of 112.

\textbf{Optimizer.} We use Adam with a learning rate of 2e-4.

\textbf{Losses.} We use the $\ell_2$ loss for quantitative comparisons as PSNR is inversely proportional to $\ell_2$. Indeed, the baselines we compare against (except TCO) all directly optimize $\ell_2$, making comparisons fairer. We generally found that $\ell_1 + \text{SSIM}$ produces more visually pleasing samples and therefore use this loss for qualitative comparisons and novel view synthesis.

\section{Dataset descriptions}

Detailed descriptions of the ShapeNet chairs and cars dataset can be found in the appendix of \citeauthor{sitzmann2019scene}.

\subsection{MugsHQ}

The MugsHQ data set was rendered with a branch of the Mitsuba Renderer~\cite{Mitsuba} adapted to import ShapeNet geometry \cite{Mitsuba_ShapeNet}.  Every scene was rendered with the same environment map (lighting conditions) and checkerboard disk platform.  ShapeNet objects were scaled by their largest bounding box dimension, centered, and placed on the platform. The object's material is a two-sided plastic designed to highlight glossy reflections and the diffuse reflectance color was randomly sampled.  For each object, 150 viewpoints were uniformly sampled over the upper hemisphere.  Each viewpoint was rendered to a $256\times256$ high dynamic range image, and then resized and tone-mapped to a linear RGB image.

\subsection{3D mountains}

We created the 3D mountains dataset by first scraping the height, latitude and longitude of the 559 highest mountains in the Alps. We then used Apple Maps to render 50 images of each mountain. Specifically, the camera was placed on a sphere of radius 600m centered on the latitude, longitude and height - 100m of the mountain. We then fixed the elevation angle to be 55 degrees (or a pitch of 35 degrees) and randomly sampled the azimuth angle between 0 and 360 degrees to capture multiple views of each mountain.

\section{Train/validation/test splits}

For each dataset we train a model and choose hyperparameters based on the lowest validation loss. All images and quantitative measurements are then made on a held out test set which is only seen after everything else has been fixed.

\subsection{Chairs}

The chairs dataset consists of 6591 scenes, with the training and validation set each having 50 views per scene and the test set having 251 views per scene, for a total of 594,267 images. The train/validation/test splits are:
\begin{itemize}
    \item Train: 4612 scenes (230,600 images)
    \item Validation: 662 scenes (33,100 images)
    \item Test: 1317 scenes (330,567 images)
\end{itemize}

\subsection{Cars}

The cars dataset consists of 3514 scenes, with the training and validation set each having 50 views per scene and the test set having 251 views per scene, for a total of 317,204 images. The train/validation/test splits are:
\begin{itemize}
    \item Train: 2458 scenes (122,900 images)
    \item Validation: 352 scenes (17,600 images)
    \item Test: 704 scenes (176,704 images)
\end{itemize}

\subsection{MugsHQ}

The MugsHQ dataset consists of 214 scenes, each with 150 views for a total of 32,100 images. The train/validation/test splits are:
\begin{itemize}
    \item Train: 186 scenes (27,900 images)
    \item Validation: 14 scenes (2,100 images)
    \item Test: 14 scenes (2,100 images)
\end{itemize}

\subsection{3D mountains}

The 3D mountains dataset consists of 559 scenes, each with 50 views for a total of 27,950 images. The train/validation/test splits are:
\begin{itemize}
    \item Train: 478 scenes (23,900 images)
    \item Validation: 26 scenes (1,300 images)
    \item Test: 55 scenes (2,750 images)
\end{itemize}

\section{Runtimes}

\subsection{Training time}

Training time for all datasets are shown in Table \ref{training-time-table}. When training on a single V100 GPU we use a batch size of 16, whereas we use a batch size of 112 when training on 8 V100s.

% Quantitative
\begin{table}[h]
\begin{center}
\begin{small}
\begin{sc}
\begin{tabular}{lcc}
\toprule
Dataset & V100 & 8 V100s \\
\midrule
Chairs & 9.7 days & 2.2 days \\
Cars & 5.5 days & 1.3 days \\
MugsHQ & 1.2 days & 6 hrs \\
Mountains & 1 day & 5 hrs \\
\bottomrule
\end{tabular}
\end{sc}
\end{small}
\caption{Training times.}
\label{training-time-table}
\end{center}
\end{table}

\subsection{Inference time}
We measured inference time with a trained model on the cars dataset running on a single Tesla V100 GPU. We took the mean and standard deviation over 1000 iterations (using 100 warmup steps).

Single image: $\mathbf{21.9\pm0.3}$ \textbf{ms}

Batch of 128 images: $\mathbf{1578.6\pm10.2}$ \textbf{ms} ($\mathbf{12.3}$ \textbf{ms} per image)

Note that for a single image this corresponds to a framerate of 45 fps, allowing for real time inference.

\section{Things that didn't work}

We experimented with several things which we found did not improve performance.

\begin{itemize}
    \item We experimented with partitioning the latent space (across channels) into a viewpoint invariant and equivariant part. We hypothesized this might help in learning complex textures and create something akin to a global texture map, but found that this did not decrease (nor increase) the loss in practice.
    \item When rotating the voxels we use trilinear interpolation to calculate the value of points that do not align with the grid. While rotations on the grid will always suffer from aliasing we hypothesized that using nearest neighbor interpolation (instead of trilinear) could help model performance. We also tried using shear rotations as these have been shown to reduce aliasing in certain cases \cite{paeth1986fast}. In practice we found that this did not make a big difference.
    \item The latent space we use in our model has shape $64 \times 32 \times 32 \times 32$. We hypothesized that increasing the spatial resolution might help improve performance. We therefore tried a latent space of size $8 \times 64 \times 64 \times 64$ but found that this performed the same as the original latent space, but was much slower to train.
\end{itemize}

\section{Samples from datasets}

We include random ground truth samples from the MugsHQ and 3D mountains dataset.

\begin{figure}[t]
\begin{center}
\centerline{\includegraphics[width=0.76\columnwidth]{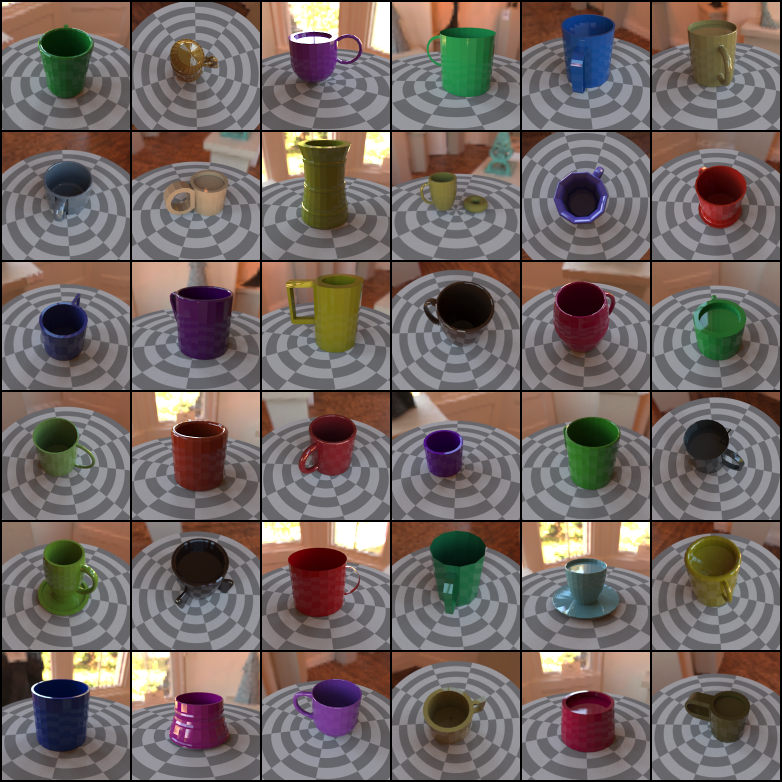}}
\caption{Random samples from the MugsHQ dataset.}
\end{center}
\end{figure}

\begin{figure}[h]
\begin{center}
\centerline{\includegraphics[width=0.76\columnwidth]{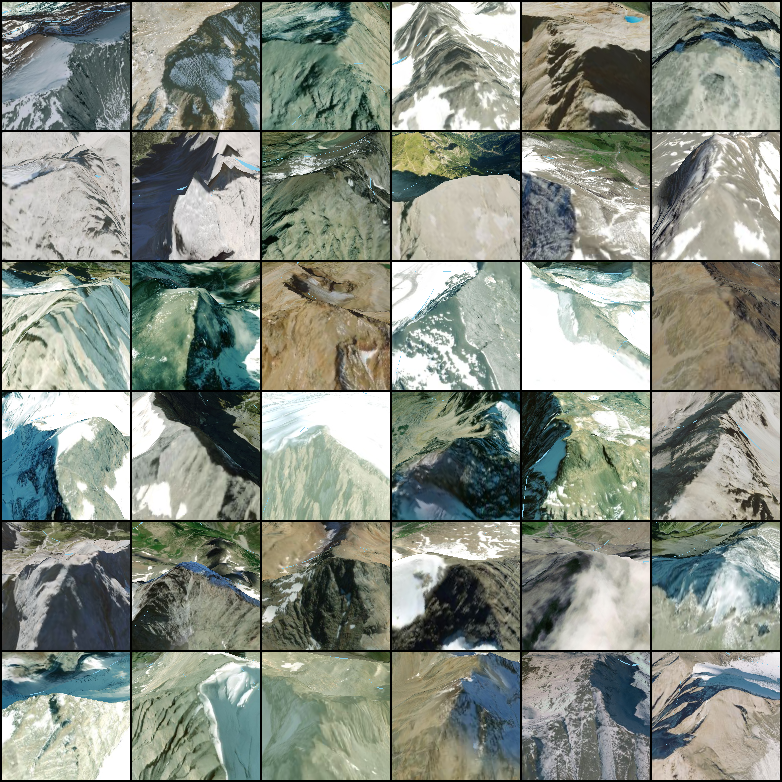}}
\caption{Random samples from the 3D mountains dataset.}
\end{center}
\end{figure}

\section{Random samples from model}

We include random novel view synthesis samples on all datasets.

\begin{figure}[t]
\small \hspace{3pt} Input \quad \hspace{5pt} Model \quad \hspace{2pt} Target \qquad \hspace{4pt} Input \quad \hspace{5pt} Model \quad \hspace{2pt} Target
\vspace{-4pt}
\begin{center}
\centerline{
    \includegraphics[width=0.159\columnwidth]{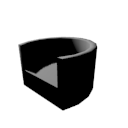}
    \includegraphics[width=0.159\columnwidth]{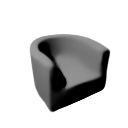}
    \includegraphics[width=0.159\columnwidth]{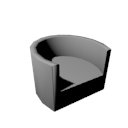}
    \hspace{5pt}
    \includegraphics[width=0.159\columnwidth]{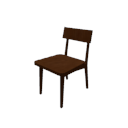}
    \includegraphics[width=0.159\columnwidth]{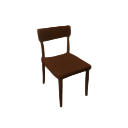}
    \includegraphics[width=0.159\columnwidth]{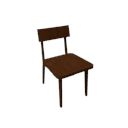}
}

\centerline{
    \includegraphics[width=0.159\columnwidth]{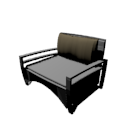}
    \includegraphics[width=0.159\columnwidth]{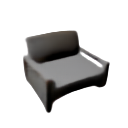}
    \includegraphics[width=0.159\columnwidth]{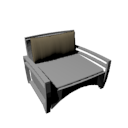}
    \hspace{5pt}
    \includegraphics[width=0.159\columnwidth]{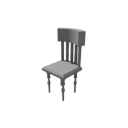}
    \includegraphics[width=0.159\columnwidth]{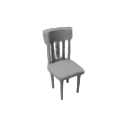}
    \includegraphics[width=0.159\columnwidth]{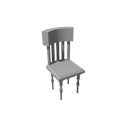}
}

\centerline{
    \includegraphics[width=0.159\columnwidth]{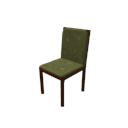}
    \includegraphics[width=0.159\columnwidth]{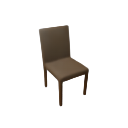}
    \includegraphics[width=0.159\columnwidth]{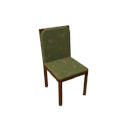}
    \hspace{5pt}
    \includegraphics[width=0.159\columnwidth]{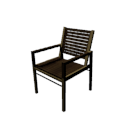}
    \includegraphics[width=0.159\columnwidth]{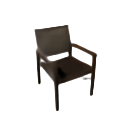}
    \includegraphics[width=0.159\columnwidth]{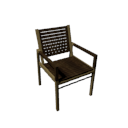}
}
\caption{Random single shot novel view synthesis samples on chairs.}
\end{center}
\end{figure}

\begin{figure}[h]
\small \hspace{3pt} Input \quad \hspace{5pt} Model \quad \hspace{2pt} Target \qquad \hspace{4pt} Input \quad \hspace{5pt} Model \quad \hspace{2pt} Target
\vspace{-4pt}
\begin{center}
\centerline{
    \includegraphics[width=0.159\columnwidth]{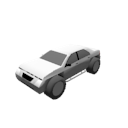}
    \includegraphics[width=0.159\columnwidth]{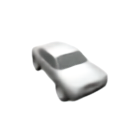}
    \includegraphics[width=0.159\columnwidth]{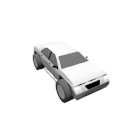}
    \hspace{5pt}
    \includegraphics[width=0.159\columnwidth]{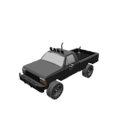}
    \includegraphics[width=0.159\columnwidth]{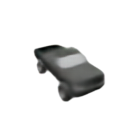}
    \includegraphics[width=0.159\columnwidth]{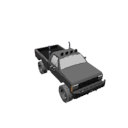}
}

\centerline{
    \includegraphics[width=0.159\columnwidth]{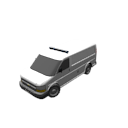}
    \includegraphics[width=0.159\columnwidth]{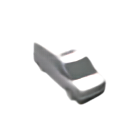}
    \includegraphics[width=0.159\columnwidth]{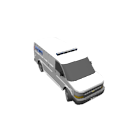}
    \hspace{5pt}
    \includegraphics[width=0.159\columnwidth]{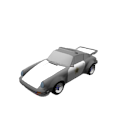}
    \includegraphics[width=0.159\columnwidth]{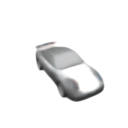}
    \includegraphics[width=0.159\columnwidth]{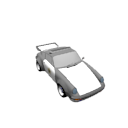}
}

\centerline{
    \includegraphics[width=0.159\columnwidth]{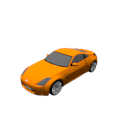}
    \includegraphics[width=0.159\columnwidth]{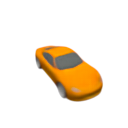}
    \includegraphics[width=0.159\columnwidth]{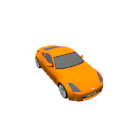}
    \hspace{5pt}
    \includegraphics[width=0.159\columnwidth]{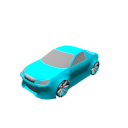}
    \includegraphics[width=0.159\columnwidth]{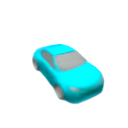}
    \includegraphics[width=0.159\columnwidth]{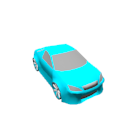}
}
\caption{Random single shot novel view synthesis samples on cars.}
\end{center}
\end{figure}

\begin{figure}[h]
\small \hspace{3pt} Input \quad \hspace{5pt} Model \quad \hspace{2pt} Target \qquad \hspace{4pt} Input \quad \hspace{5pt} Model \quad \hspace{2pt} Target
\vspace{-4pt}
\begin{center}
\centerline{
    \includegraphics[width=0.159\columnwidth]{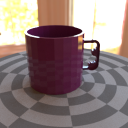}
    \includegraphics[width=0.159\columnwidth]{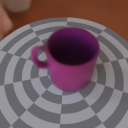}
    \includegraphics[width=0.159\columnwidth]{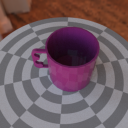}
    \hspace{5pt}
    \includegraphics[width=0.159\columnwidth]{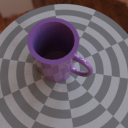}
    \includegraphics[width=0.159\columnwidth]{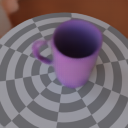}
    \includegraphics[width=0.159\columnwidth]{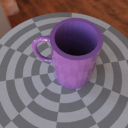}
}

\centerline{
    \includegraphics[width=0.159\columnwidth]{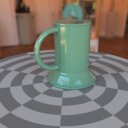}
    \includegraphics[width=0.159\columnwidth]{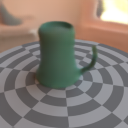}
    \includegraphics[width=0.159\columnwidth]{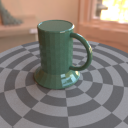}
    \hspace{5pt}
    \includegraphics[width=0.159\columnwidth]{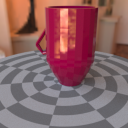}
    \includegraphics[width=0.159\columnwidth]{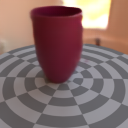}
    \includegraphics[width=0.159\columnwidth]{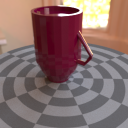}
}

\centerline{
    \includegraphics[width=0.159\columnwidth]{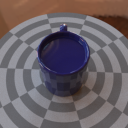}
    \includegraphics[width=0.159\columnwidth]{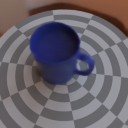}
    \includegraphics[width=0.159\columnwidth]{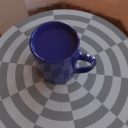}
    \hspace{5pt}
    \includegraphics[width=0.159\columnwidth]{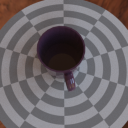}
    \includegraphics[width=0.159\columnwidth]{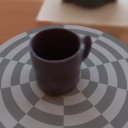}
    \includegraphics[width=0.159\columnwidth]{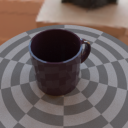}
}
\caption{Random single shot novel view synthesis samples on MugsHQ.}
\end{center}
\end{figure}

\begin{figure}[h]
\small \hspace{3pt} Input \quad \hspace{5pt} Model \quad \hspace{2pt} Target \qquad \hspace{4pt} Input \quad \hspace{5pt} Model \quad \hspace{2pt} Target
\vspace{-4pt}
\begin{center}
\centerline{
    \includegraphics[width=0.159\columnwidth]{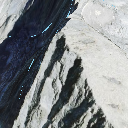}
    \includegraphics[width=0.159\columnwidth]{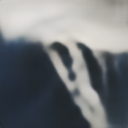}
    \includegraphics[width=0.159\columnwidth]{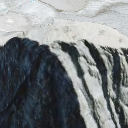}
    \hspace{5pt}
    \includegraphics[width=0.159\columnwidth]{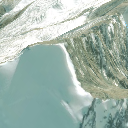}
    \includegraphics[width=0.159\columnwidth]{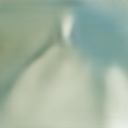}
    \includegraphics[width=0.159\columnwidth]{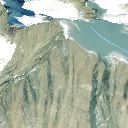}
}

\centerline{
    \includegraphics[width=0.159\columnwidth]{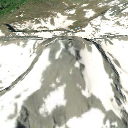}
    \includegraphics[width=0.159\columnwidth]{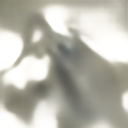}
    \includegraphics[width=0.159\columnwidth]{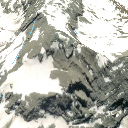}
    \hspace{5pt}
    \includegraphics[width=0.159\columnwidth]{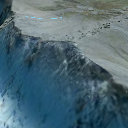}
    \includegraphics[width=0.159\columnwidth]{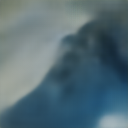}
    \includegraphics[width=0.159\columnwidth]{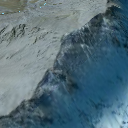}
}

\centerline{
    \includegraphics[width=0.159\columnwidth]{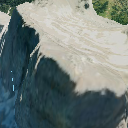}
    \includegraphics[width=0.159\columnwidth]{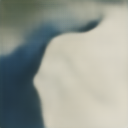}
    \includegraphics[width=0.159\columnwidth]{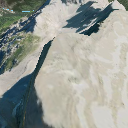}
    \hspace{5pt}
    \includegraphics[width=0.159\columnwidth]{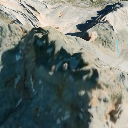}
    \includegraphics[width=0.159\columnwidth]{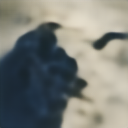}
    \includegraphics[width=0.159\columnwidth]{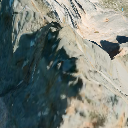}
}
\caption{Random single shot novel view synthesis samples on 3D mountains.}
\end{center}
\end{figure}

%\begin{figure}[h]
%\begin{center}

%\centerline{
%    \raisebox{1.25pt}{\includegraphics[width=0.159\columnwidth]{figures/mugs-input-5.png}}%
%    \hspace{3pt}%
%    \textcolor{darkgray1}{\vrule width 1pt height 35pt depth -4pt}%
%    \hspace{2pt}%
%    \includegraphics[width=0.83\columnwidth]{figures/mugs-novel-5.png}
%}

%\centerline{
%    \raisebox{1.25pt}{\includegraphics[width=0.159\columnwidth]{figures/mugs-input-4.png}}%
%    \hspace{3pt}%
%    \textcolor{darkgray1}{\vrule width 1pt height 35pt depth -4pt}%
%    \hspace{2pt}%
%    \includegraphics[width=0.83\columnwidth]{figures/mugs-novel-4.png}
%}

%\centerline{
%    \raisebox{1.25pt}{\includegraphics[width=0.159\columnwidth]{figures/mugs-input-7.png}}%
%    \hspace{3pt}%
%    \textcolor{darkgray1}{\vrule width 1pt height 35pt depth -4pt}%
%    \hspace{2pt}%
%    \includegraphics[width=0.83\columnwidth]{figures/mugs-novel-7.png}
%}

%\caption{Additional novel view synthesis examples for MugsHQ.}
%\end{center}
%\end{figure}

%\fi

\end{document}